\documentclass[10pt,twocolumn,letterpaper]{article}

\usepackage{iccv}              

\usepackage{algorithm}
\usepackage{algpseudocode}
\usepackage{amsmath}
\usepackage{multirow}
\usepackage{makecell}

\definecolor{iccvblue}{rgb}{0.21,0.49,0.74}
\usepackage[pagebackref,breaklinks,colorlinks,allcolors=iccvblue]{hyperref}

\def\paperID{6492}
\def\confName{ICCV}
\def\confYear{2025}

\title{Contact-Aware Amodal Completion for Human-Object Interaction via Multi-Regional Inpainting}

\author{$\text{Seunggeun Chi}^1$\\
Purdue University\\
West Lafayette, IN, USA\\
{\tt\small chi65@purdue.edu}
\and
Enna Sachdeva,\quad Pin-Hao Huang,\quad Kwonjoon Lee\\
Honda Research Institute USA\\
San Jose, CA, USA\\
{\tt\small \{enna\_sachdeva, pin-hao\_huang, kwonjoon\_lee\}@honda-ri.com}
}

\begin{document}
\twocolumn[{%
\renewcommand\twocolumn[1][]{#1}%
\maketitle

\vspace{-2em}
\begin{center}
    \centering
    \captionsetup{type=figure}
    \includegraphics[width=1.00\textwidth]{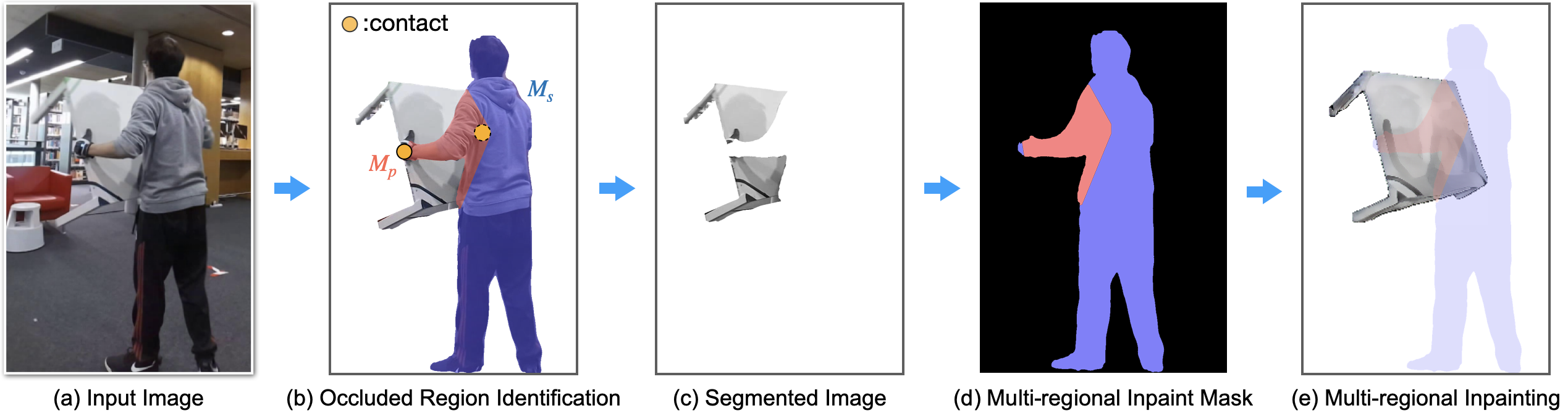}
    \vspace{-1.5em}
    \captionof{figure}{Our amodal completion pipeline for human-object interaction. (a) Occlusions frequently occur during human-object interactions. (b) By applying a convex hull based method to contact points (shown with yellow circles), we identify primary region \textcolor{RedOrange}{$M_p$} highly likely to contain occluded parts, as well as secondary region \textcolor{Violet}{$M_s$} that exhibit a lower, yet present, probability of occlusion. (e) Our multi-regional inpainting method completes the segmented image using these masks, without additional training.}
\label{fig:teaser}
\end{center}%
}]

\setcounter{footnote}{1}
\footnotetext{Work done at Honda Research Institute} 

\begin{abstract}
Amodal completion, the task of inferring the complete appearance of objects despite partial occlusions, is crucial for understanding complex human–object interactions (HOI) in computer vision and robotics. Existing methods, including pre-trained diffusion models, often struggle to generate plausible completions in dynamic scenarios due to their limited understanding of HOI. To address this challenge, we propose a novel approach that leverages physical prior knowledge alongside a specialized multi-regional inpainting technique tailored for HOI. By incorporating physical constraints derived from human topology and contact information, we define two distinct regions: the primary region, where occluded object parts are most likely to reside, and the secondary region, where occlusions are less probable. Our multi-regional inpainting method employs customized denoising strategies across these regions within a diffusion model, thereby enhancing the accuracy and realism of generated completions in both shape and visual detail. Experimental results demonstrate that our approach substantially outperforms existing methods in HOI scenarios, advancing machine perception toward a more human-like understanding of dynamic environments. Furthermore, we show that our pipeline remains robust even without ground-truth contact annotations, broadening its applicability to tasks such as 3D reconstruction and novel view/pose synthesis. 
\vspace{-1em}
\end{abstract}    
\section{Introduction}

Understanding human-object interactions (HOI) is a fundamental challenge in the fields of computer vision and robotics. Accurate perception of these interactions enables a wide range of applications, from autonomous robots that can safely navigate human environments to augmented reality systems that seamlessly integrate virtual objects into the real world. However, a significant obstacle in interpreting these interactions is the presence of occlusions, where parts of objects or humans are hidden from view due to overlapping elements in the scene.

Amodal completion~\cite{chen2016amodal, emmanouil2014amodal} offers a promising solution to this problem by enabling systems to infer the complete shape and extent of partially occluded objects. This cognitive ability, inherent in human perception, allows us to recognize objects and estimate occluded parts of human body even when we cannot see them in their entirety.

Recently, pre-trained diffusion models~\cite{rombach2022high} have emerged as powerful tools for amodal completion due to their generative capabilities. These models can generate plausible completions of occluded regions, enhancing the overall understanding of complex scenes~\cite{ozguroglu2024pix2gestalt,xu2024amodal,yildirim2023inst}.
A straightforward approach is to apply inpainting/outpainting on the segment of the occluder. However, applying diffusion models directly to occluded images without proper regioning often leads to implausible or incorrect completions. For HOI, when an object is occluded by human, the occluder region is often significantly larger than the actual occluded area of the object as shown in \Cref{fig:teaser}.
Inaccurately identifying occluded region causes diffusion models to generate overextended or inaccurate completions, as the inpainting process affects a larger area than necessary.

To address this issue, we introduce a novel \textit{region identification} method that precisely defines the areas requiring inpainting. Specifically, we divide the occluded region into two distinct areas: a primary region and a secondary region. The primary region, which is more likely to contain the occluded parts of the object, is identified using a \textit{contact-aware convex hull} operation that incorporates contact information and the human-object boundary. This targeted approach focuses the inpainting process on the most relevant area, improving the accuracy and plausibility of the completions. In contrast, the secondary region encompasses the remaining parts of the occluder, with a lower probability of containing occluded object details. By distinguishing these two regions, we apply inpainting more effectively, reducing unnecessary alterations in areas unlikely to contain occluded information.

Building on this region identification, we introduce a novel inpainting method,\textit{ multi-regional inpainting}, which operates without requiring additional training. This method applies differentiated denoising strategies across the regions: it constructs a coarse structure in the primary region and then adds finer details in the secondary region. By implementing these multi-regional denoising strategies, we enhance the model's capacity to produce accurate completions in the primary region while maintaining the integrity of the secondary region.
With these amodally completed images, we demonstrate that the visually enriched images can boost applications such as 3D human and object reconstruction with Gaussian Splatting~\cite{kerbl20233d} on HOI. 

In summary, our contributions are:
\begin{itemize} 
    \item \textbf{Amodal Completion Framework for Human Object Interaction:} To the best of our knowledge, our work is the first to address amodal completion in HOI. We develop a framework that accurately predicts the complete appearance of both the human and the object during interaction. By leveraging distinct constraints inherent in HOI, our approach precisely identifies occluded regions.

    \item \textbf{Multi-Regional Inpainting Method:} We introduce a novel inpainting technique that extends the pre-trained diffusion model~\cite{rombach2022high} without requiring additional training. This method employs differentiated denoising strategies across regions with different levels of priority, enabling more precise completion.
    
    \item \textbf{Applications of Amodal Completion:} For practical applicability, we propose a pipeline that operates without ground-truth contact information. In addition, we demonstrate that our amodal completion method for HOI supports various applications, including 3D reconstruction with Gaussian Splatting and novel-view/pose synthesis for humans and objects.

\end{itemize}

\section{Related Work}

\subsection{Amodal Segmentation and Completion}
Amodal segmentation and completion address the challenge of reconstructing fully visible object shapes from partially occluded views, enhancing scene comprehension. Early approaches, such as the bilayer convolutional network by \cite{ke2021deep}, improve segmentation accuracy by differentiating occluders from occludees, while variational autoencoders~\cite{ling2020variational} model latent structures for plausible occlusion completion. Bayesian generative models~\cite{sun2022amodal} and vector-quantized representations~\cite{gao2023coarse} introduce probabilistic and coarse-to-fine methods for handling various occlusion levels. To capture mutual occlusions in structured scenes, \cite{zhang2024amodal} proposed a holistic relation inference framework.

Recent advancements in amodal completion include diffusion-based models, such as \cite{xu2024amodal} and Pix2gestalt~\cite{ozguroglu2024pix2gestalt}, which leverage segmentation order analysis and synthetic whole-part pairs to accurately infer occluded areas. Self-supervised methods~\cite{zhan2020self} allow models to infer occlusion relationships, while new datasets with 3D ground truth~\cite{zhan2024amodal} provide valuable benchmarks for real-world scenarios. In contrast to these approaches that restrict the inpainting region to a single mask or operate without one, our method handles multiple inpainting regions with varying priorities.

\begin{figure*}[!t]
\centering
\begin{center}
    \centering
    \captionsetup{type=figure}
    \includegraphics[width=0.98\textwidth]{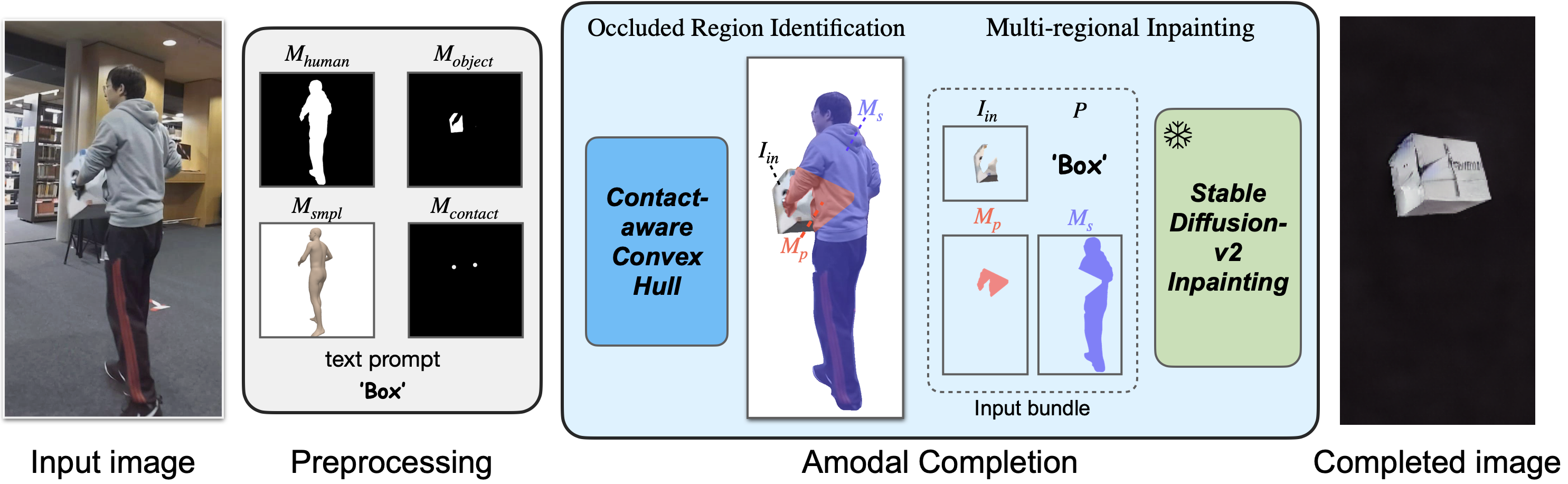}
    \captionof{figure}{The overall pipeline of our proposed method. Given an RGB image of human-object interaction, our pipeline utilizes human, object, contact, SMPL mask information, represented by $M_{human}$, $M_{object}$, $M_{contact}$, $M_{smpl}$, respectively, and a text prompt $\mathcal{P}$ describing the object category. Firstly, it leverages $M_{human}$, $M_{object}$ and $M_{contact}$ to identify key regions of interests: primary \textcolor{RedOrange}{$M_p$} and secondary \textcolor{Violet}{$M_s$} occluded region on the occluder. The identified regions \textcolor{RedOrange}{$M_p$} and \textcolor{Violet}{$M_s$}, text prompt $\mathcal{P}$ along with segmented object image $I_\text{in}$, are then utilized for the amodal completion, a process where both the human and object can interchangeably act as an occluder or an occludee.}
    \vspace{-1em}
    \label{fig:pipeline}
\end{center}%
\end{figure*}

\subsection{Human-Object Interaction and Occlusion}
Human-Object Interaction (HOI) research often faces occlusion challenges, obscuring key parts of human-object interactions. Contact estimation methods, such as CONTHO~\cite{nam2024joint}, HOT~\cite{chen2023detecting}, and DECO~\cite{tripathi2023deco}, help predict occluded regions by identifying interaction points, preserving dynamics in both 2D and 3D views. Models like LEMON~\cite{yang2024lemon} and COMA~\cite{coma} enhance scene understanding by capturing spatial relationships and affordance cues. Additionally, 3D reconstruction methods like CHORE~\cite{xie2022chore}, VisTracker~\cite{xie2023vistracker}, and HDM~\cite{xie2023template_free} improve pose estimation from partial views, collectively supporting a more complete understanding of HOI under occlusion.




\section{Preliminary}\label{section:preliminary}
\subsection{Convex Hull}
A convex hull is a fundamental geometric concept in computational geometry and computer vision, frequently employed to delineate regions of interest, infer spatial relationships, and establish bounding areas for subsequent analysis~\cite{jayaram2016convex,karavelas2013convex,rosin2000shape,sirakov2004search,yang2013graph,wang2014convex}. Its ability to simplify complex shapes and accurately approximate object boundaries significantly enhances the efficiency of spatial analyses in image processing and pattern recognition tasks.

The convex hull represents the smallest convex set that contains all given points in a 2D space. Given a set of points:
\vspace{-1em}
\begin{flalign}
    C = \{p_1, p_2, \dots, p_n\} &
\end{flalign}
where each point $p_i = (x_i, y_i) \in \mathbb{R}^2$ defines a location in the plane, the convex hull \(H\) of the set \(C\) is the smallest convex polygon that encloses all the points in \(C\). Formally, we denote the convex hull as:
\begin{flalign}
    H = \text{ConvexHull}(C), &
\end{flalign}
where \(\text{Hull}(C)\) is the boundary formed by connecting the outermost points in \(C\) such that every point lies either on this boundary or within the polygon. This polygon can be visualized as a ``tight rubber band" stretched around the outermost points. To create a mask \(M_{hull}\) representing the convex hull, we define it as follows:
\begin{flalign}\label{eq:hull_mask}
    M_{hull}(x, y) = 
    \begin{cases} 
        1 & \text{if } (x, y) \in H \\
        0 & \text{otherwise}
    \end{cases} &
\end{flalign}
This binary mask \(M_{hull}\) assigns a value of 1 to pixels inside the convex hull, representing the enclosed area, and a value of 0 to all other pixels.
\section{Method}
As shown in \cref{fig:pipeline}, we address the amodal completion problem in human-object interactions by leveraging distinctive characteristics inherent to dynamic scenarios. Unlike typical occlusions observed in static scenes, human-object interactions present specific challenges and unique features: (1) the visible regions of subjects often exhibit concave shapes or multiple segmented parts, (2) human body topology is accessible, and (3) the presence of human-object contact points provides crucial spatial relationship cues.

Motivated by our observations of concave and segmented appearances (1), we employ the convex hull operation to effectively identify regions requiring completion. Additionally, by utilizing the topology of the human body (2), we can accurately confine body part locations enabling estimation of human-object contact points (3). Integrating this contact information with the convex hull expands and refines the region targeted for amodal completion.

Based on these insights, we propose a pipeline consisting of two main components: Occluded Region Identification (\cref{sec:region_identification}) and Multi-Regional Inpainting (\cref{sec:multi_region}). To facilitate practical application in in-the-wild scenarios, we further introduce a method for estimating human-object contact information without relying on ground-truth annotations, detailed in \cref{sec:in-the-wild}.

\begin{figure*}[t]
\centering
\includegraphics[width=0.98\linewidth]{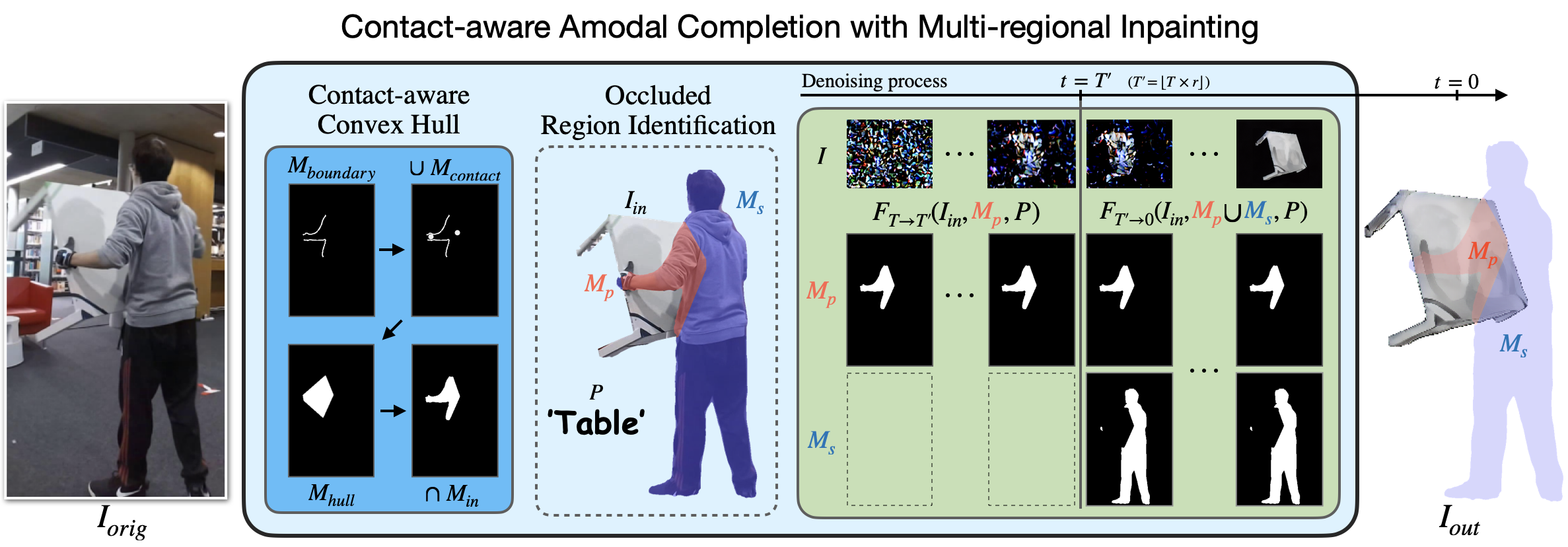}
\vspace{-0.5em}
\caption{Given an RGB image ($I_{orig}$) of human-object interaction, we obtain $M_{bounday}$ using dilation operation to the mutually exclusive mask of human $M_{human}$ and object $M_{object}$. This information along with contact $M_{contact}$ is used to obtain convex hull $M_{hull}$, which then yields $M_p$. These identified occluded region masks \textcolor{RedOrange}{$M_p$} and \textcolor{Violet}{$M_s$} are used for multi-regional inpainting for amodal completion. During the denoising process, the strength parameter \(r\) regulates the initiation timestep for inpainting the secondary region.}
\vspace{-0.5em}
\label{fig:object_completion}
\end{figure*}

\subsection{Problem Formulation} \label{sec:problem_definition}
Similar to the setup in \cite{xu2024amodal}, we define the amodal completion problem as:
\begin{equation}
    I_\text{out} = F_{s \rightarrow e}(I_\text{in}, M_\text{in}, \mathcal{P}),
\end{equation}
where \( I_\text{in} \in \mathbb{R}^{H \times W \times 3} \) is the segmented input image containing only the visible parts of the subject, \( M_\text{in} \in \{0,1\}^{H \times W} \) is the input mask confining the area of interest for inpainting, and $\mathcal{P}$ is a text prompt providing contextual guidance for completion. The function \( F_{s \rightarrow e} \) represents the diffusion denoising process, which reconstructs the occluded region within the mask \( M_\text{in} \), operating from the starting step \( s \) to the ending step \( e \), and outputs the completed image \( I_\text{out} \in \mathbb{R}^{H \times W \times 3} \).
To describe the inpainting process in multiple stages, we introduce a new notation \( | \), which allows us to decompose the process into intermediate steps. For any intermediate timestep \( e < t < s \), the inpainting process can be broken down as follows:
\begin{flalign}
    I_\text{out} &=F_{s \rightarrow e}(I_\text{in}, M_\text{in}, \mathcal{P}) \nonumber\\
    &= F_{s \rightarrow t}(I_\text{in}, M_\text{in}, \mathcal{P}) | F_{t \rightarrow e}(I_\text{in}, M_\text{in}, \mathcal{P}).\label{eq:decompose}
\end{flalign}
This formulation allows flexibility in representing the inpainting process at various stages, facilitating controlled inpainting with varying levels of completion across different regions within the mask.
\vspace{-1em}
\paragraph{Stable Diffusion with a Strength Parameter}
For the diffusion model \( F \), we utilize the pre-trained Stable Diffusion-v2 inpainting model (SD-inpaint)~\cite{rombach2022high}. The SD-inpaint model includes a strength parameter \( r \), which modulates the amount of noise applied within the inpainting mask. This parameter \( r \), ranging from 0 to 1, controls the intensity of noise added to the masked region. At \( r\!=\!1 \), the model begins denoising from pure noise, fully overwriting the initial image in the masked area. Conversely, as \( r \) approaches 0, less noise is introduced, preserving more of the original image information in the masked region.
The SD-inpaint process with the strength parameter \( r \) can be expressed as:
\begin{flalign}
I_\text{out} &= F_{T\rightarrow 0}(I_\text{in}, M_\text{in}, \mathcal{P}, r) \nonumber\\
             &= F_{T'\rightarrow 0}(I_\text{in}, M_\text{in}, \mathcal{P}), \text{where } T' = \lfloor T \cdot r \rfloor \label{eq:strength}
\end{flalign}
Here, \( T=50 \) represents the total number of diffusion timesteps of DDIM scheduler~\cite{song2020denoising}, and \( \lfloor \cdot \rfloor \) indicates rounding down to the nearest integer.

\subsection{Multi-Regional Inpainting with Convex Hull}\label{sec:amodal_completion}

Diffusion models are known to establish coarse structures in the initial stages of the denoising process, gradually refining details as the process advances. Building on this characteristic, and drawing inspiration from recent mask-inpainting strategies~\cite{coma, xu2024amodal, li2024genzi} that effectively restrict the inpainting area, we propose a novel multi-regional mask-inpainting approach. This approach enhances the diffusion model's effectiveness by confining the completion area using physical constraints, particularly focusing on the contact points between the human and object within the scene.


\begin{figure*}[t!]
\includegraphics[width=\linewidth]{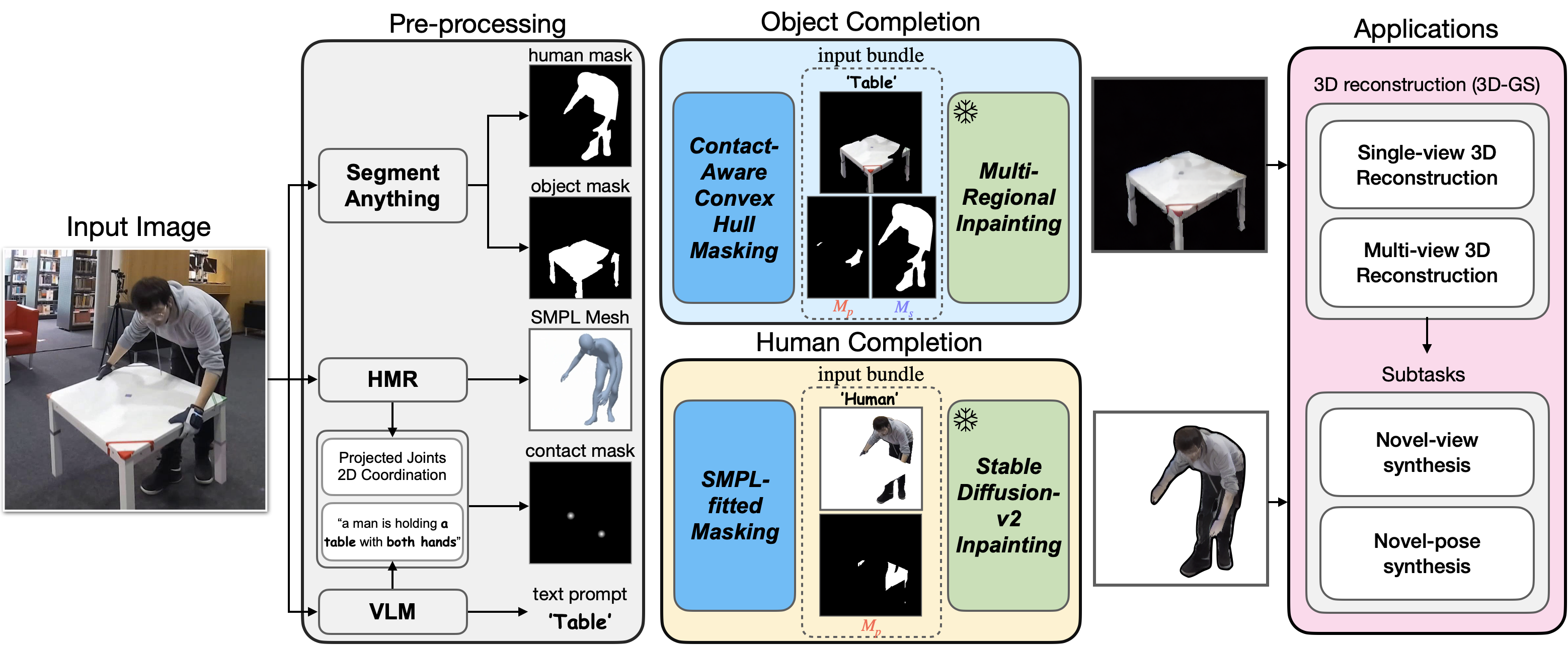}
\captionof{figure}{Our amodal completion pipeline designed to process in-the-wild data without relying on ground-truth contact annotations.}
\label{fig:supple_pipeline}
\end{figure*}

\subsubsection{Occluded Region Identification}
\label{sec:region_identification}

We introduce a method to identify occluded regions with different levels of priority, improving inpainting accuracy by focusing on areas with a high likelihood of occlusion derived from contact points. In the occluded region identification stage, we generate an input mask tuple \( \{M_p, M_s\} \) using a convex hull operation as introduced in \cref{section:preliminary}. 
\vspace{-1em}
\paragraph{Contact-aware Convex Hull}
Our contact-aware convex hull process, illustrated in \cref{fig:object_completion}, refines the inpainting region by incorporating proximity and interaction cues. First, we compute an occlusion boundary mask \( M_{\text{boundary}} \) by applying a dilation operation~\cite{serra1983image} to the mutually exclusive masks of the human (\( M_{\text{human}} \)) and object (\( M_{\text{obj}} \)), which segment the visible parts of the human and object in the image. This step highlights areas where the human and object are in close proximity, marking potential occlusions.

We then define a set of points \( C \) by combining \( M_{\text{boundary}} \) with a binary contact map \( M_{\text{contact}} \), resulting in \( C = M_{\text{boundary}} \cup M_{\text{contact}} \). From this combined set \( C \), we compute the convex hull \( H = \text{Hull}(C) \), forming the smallest convex polygon enclosing all points in \( C \), as described in \cref{section:preliminary}. Then we assign values to the convex hull \( M_{\text{hull}} \) with~\cref{eq:hull_mask}. The contact-aware convex hull mask, named as primary mask \( M_{p} \), is then derived by intersecting the occluder mask \( M_{\text{in}} \) with \( M_{\text{hull}} \), yielding \( M_{p} = M_{\text{in}} \cap M_{\text{hull}} \). This mask excludes visible parts of the occludee and designates the primary region where occlusion is most likely, effectively confining the inpainting area to enhance completion quality.
While \( M_{p} \) captures most of the occluded areas, as shown in \(I_{out}\) of \cref{fig:object_completion} , it may not cover all occluded regions. Remaining areas within \( M_{\text{in}} \) that are outside \( M_{p} \) are referred to as the secondary region \( M_{s} = M_{\text{in}} \setminus M_{p} \). These secondary regions represent areas that still need handling to ensure comprehensive coverage in the inpainting process.

\subsubsection{Multi-Regional Inpainting}\label{sec:multi_region}
As outlined in \cref{sec:region_identification}, we identify two key regions: the primary region \( M_p \), where occluded areas highly likely exists, and the secondary region \( M_s \), which may require further inpainting refinement. These regions form the foundation of our multi-regional inpainting method, designed to adaptively address varying occlusion levels within a unified framework. We extend the SD-inpaint pipeline to handle the multi-regional masking by adapting the expressions in~\cref{eq:decompose,eq:strength}:
\begin{flalign}
I_\text{out} &= F_{T \rightarrow 0}(I_\text{in}, \{M_p, M_s\}, \mathcal{P}, r)\\
&:= F_{T \rightarrow T'}(I_\text{in}, M_p, \mathcal{P}) \,|\, F_{T' \rightarrow 0}(I_\text{in}, M_p \small{\cup} M_s, \mathcal{P}),
\end{flalign}
where  $T' = \lfloor T \cdot r \rfloor$. This formulation enables an adaptive inpainting process that first establishes the coarse structure within \(M_p\) and then progressively refines details across both \(M_p\) and \(M_s\), guided by the initial structure in the primary region. This multi-regional approach ensures seamless blending and alignment between regions. As illustrated in \cref{fig:object_completion}, the parameter \(r\) controls the inpainting strength applied to the secondary region. Visually, \(r\) adjusts the horizontal placement of the vertical bar in green box, thereby influencing when inpainting of the secondary region begins.


Unlike existing diffusion-based inpainting algorithms~\cite{rombach2022high,xu2024amodal} that typically handle a single input mask, our method is specifically designed to manage multiple regions simultaneously, applying different noise intensities and strategies for each. This multi-regional inpainting process leverages adaptive strengths to prioritize occluded areas near the occlusion boundaries while refining potential regions, all within a single framework and without additional training. This enables superior coverage and accuracy for dynamic occlusion scenarios, a clear advantage over traditional inpainting techniques.

\begin{figure*}[t!]
    \centering
    \captionsetup{type=figure}
    \includegraphics[width=1.00\textwidth]{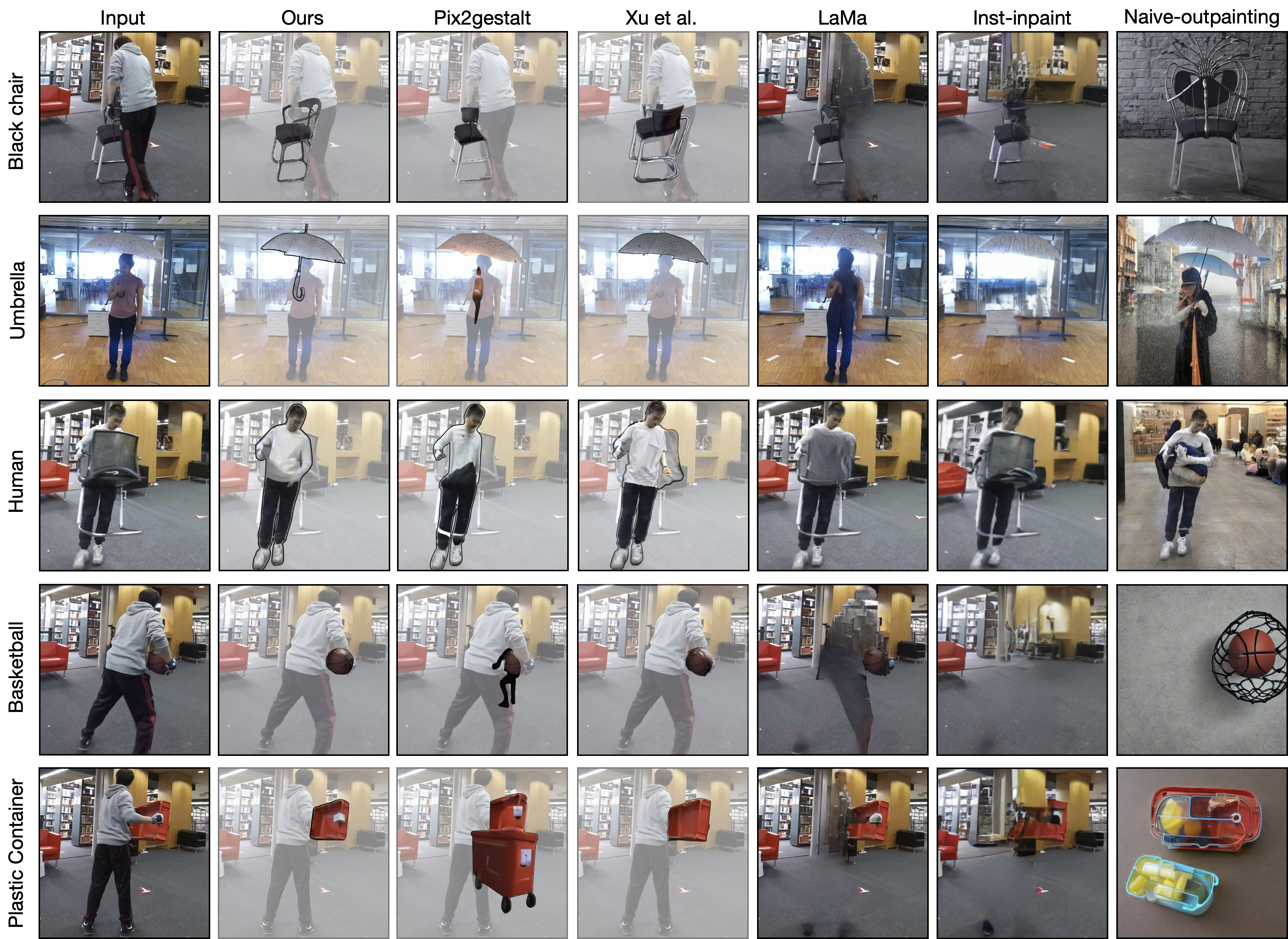}
    \captionof{figure}{\vspace{-0.7em}Qualitative comparison. Our approach produces more accurate and realistic results through effective region identification.}
    \vspace{-0.7em}
    \label{fig:qualitative}
\end{figure*}

\subsection{Amodal Completion on In-the-Wild Data}\label{sec:in-the-wild}
To extend our approach to real-world data, we propose a method for generating the necessary inputs for our pipeline without relying on ground truth annotations.
As visulaized in \cref{fig:supple_pipeline}, instead of requiring ground-truth 3D meshes, segmentation masks, contact information, and object categories, we employ Segment Anything (SAM)~\cite{ravi2024sam} to generate human and object masks, Human Mesh Recovery (HMR) models~\cite{baradel2024multi} to estimate SMPL parameters for the human body, and VLM~\cite{openai2024chatgpt4} to produce a single-sentence description of the interaction between the human and object.
For example, as shown in~\cref{fig:supple_pipeline}, a prompt-engineered VLM takes an image and outputs both a textual description “a man is holding an object with both hands” and the corresponding SMPL joint IDs (22 and 23 for “both hands").
With SMPL regressor, we can identify the 3D coordination of "both hands", and then we generate contact mask by projecting them into 2D space.
This streamlined pipeline enables amodal completion on in-the-wild data, making it practical for real-world applications.
We also propose a regioning for human amodal completion utilizing SMPL parameters; details are in the supplementary.

\section{Experiments}
\subsection{Datasets \& Evaluation Metrics}
\paragraph{BEHAVE~\cite{bhatnagar2022behave}} includes 321 RGB-D sequences of human-object interactions, featuring 8 subjects with 20 objects in indoor settings, captured by 4 Kinect cameras. It provides 3D SMPL and object fits with annotated contacts. Among 4,500 testing frames, we filter images with occlusion ratio ($10\sim70\%$), resulting in 1,709 test images.
\noindent\paragraph{InterCap~\cite{huang2022intercap}} includes 223 RGB-D videos of human object interactions, captured from 6 views with 10 subjects and 10 objects. Using 1 fps sampling and occlusion-based filtering, we obtain 1,034 test images.
\noindent\vspace{-1.3em}\paragraph{Evalutaion metrics} 
In the experiment section, we report only the results for object amodal completion. CLIP~\cite{radford2021learning} score and mIoU is used for evaluation metrics following~\cite{xu2024amodal,ozguroglu2024pix2gestalt}.  The CLIP score measures alignment between generated images and object category prompts, while mIoU assesses overlap between predicted and groundtruth amodal masks. We calculate the CLIP score after the segmentation with SAM~\cite{ravi2024sam}.
3D reconstruction performance is evaluated using Chamfer distance between predicted and GT human/object meshes. We also report a win-rate derived from 1-on-1 user preference studies against other baseline methods.

\subsection{Amodal Completion Results}
In the following subsections, we present the amodal completion results from our pipeline. Unless otherwise noted, \textbf{Ours} refers to the in-the-wild pipeline as described in \cref{sec:in-the-wild}, focused solely on objects with $r=0.5$.

In~\cref{tab:compare_amodal_completion}, we compare our method against baselines such as pix2gestalt~\cite{ozguroglu2024pix2gestalt}, LaMa~\cite{suvorov2022resolution}, Inst-inpaint~\cite{yildirim2023inst}, and Naive outpainting~\cite{rombach2022high}, 
demonstrating superior performance across both CLIP score and mIoU metrics. Our approach consistently achieves the highest scores in mIoU, surpassing competing methods in generating amodal completions that accurately capture occluded regions. These results highlight our model's effectiveness in producing contextually aligned and precise amodal completions. However, in terms of CLIP score, the Naive outpainting method achieves the highest performance. This is because Naive outpainting generates content across the entire canvas, inherently favoring broader visual alignment with the query.

We visualize our result in~\cref{fig:qualitative}, demonstrating that our proposed method effectively confines the inpainting region with contact information, resultingly completes the occluded object and human with accurate shape as well as the appearance.
More qualitative results and the human amodal completion results and can be found in the supplementary.

\begin{table}[t]
\centering

\resizebox{0.9\linewidth}{!}{
\begin{tabular}{l c c c c c}
\hline
\multirow{2}{*}{\textbf{Method}} & \multicolumn{2}{c}
{\textbf{BEHAVE}} & \multicolumn{2}{c}{\textbf{InterCap}} & \multirow{2}{*}{\textbf{Win-rate}} \\
\cline{2-5}
& CLIP & mIoU & CLIP & mIoU \\
\hline
\hline
Naive outpainting~\cite{rombach2022high} & \textbf{27.34} & 50.92\% & \textbf{27.55} & 52.07\% & 94.0\%\\
LaMa~\cite{suvorov2022resolution} & 25.97 & 60.47\% & 26.43 & 51.38\% & 92.4\%\\
Inst-Inpaint~\cite{yildirim2023inst} & 26.08 & 63.71\% & 26.12 & 57.54\% & 88.0\% \\
pix2gestalt~\cite{ozguroglu2024pix2gestalt} & 23.45 & 69.58\% & 26.14 & 68.32\% & 68.0\%\\
Xu et al.~\cite{xu2024amodal} & 26.34 & 71.03\% & 26.21 & 69.23\% & 65.8\%\\
\textbf{Ours} & 26.91 & \textbf{77.64}\% & 26.97 & \textbf{72.34}\% & - \\
\hline
\end{tabular}
}
\vspace{-0.5em}
\caption{Comparison of amodal completion performance with baseline models. Our method achieves the highest mIoU by identifying occluded regions. \textbf{Win-rate} indicates the ratio of user preferences for our method compared to each baseline in user studies.}
\label{tab:compare_amodal_completion}
\end{table}

\begin{table}[t]
\centering
\resizebox{1.0 \linewidth}{!}{
\begin{tabular}{c l c c c c}
\hline
 &\textbf{Method} & $r$ & Region & CLIP $\uparrow$ & mIoU $\uparrow$ \\
\hline
\hline
- & Input image & - & - & 21.75 & 34.43\% \\
\hline
\multirow{4}{*}{\rotatebox[origin=c]{90}{Single}} & Naive outpainting & - & ${I_\text{in}}^{\complement}$ & 27.34 & 50.92\%  \\
& Human mask & $r\!=\!1.0$ & $M_p\small{\cup} M_s$ & 26.27 & 69.98\%  \\
& Convex hull w/o contact & $r\!=\!0.0$ & $M_p$ & 26.43  & 75.24\% \\
& Convex hull w/ contact & $r\!=\!0.0$ & $M_p$ & 26.63  & 76.11\% \\
\hline
\multirow{2}{*}{\rotatebox[origin=c]{90}{Multi}} & \textbf{Ours} & $r\!=\!0.5$ & $\{M_p, M_s\}$ & 26.91 & 77.64\% \\
& Ours w/ GT-contact & $r\!=\!0.5$ & $\{M_p, M_s\}$ & 27.07  & 80.15\% \\
\hline
\end{tabular}
}
\vspace{-0.5em}
\caption{Ablation study comparing single-region and multi-region strategies for Amodal Completion on the BEHAVE dataset. Our multi-regional approach outperforms single-region methods.}
\label{tab:mask_inpaint_strategy}
\end{table}

\subsection{Ablation Study}
\paragraph{Effect of Different Mask Inpainting Strategies} We present an ablation study on various regioning strategies for amodal completion on the BEHAVE dataset in \cref{tab:mask_inpaint_strategy}.
The results demonstrate that straightforward single mask approaches such as naive outpainting and human mask approaches inadequately capture occluded regions, leading to unrealistic reconstructions.
In contrast, our proposed contact-aware multi-regional inpainting strategy effectively leverages spatial consistency from human-object interactions, significantly improving accuracy and realism.
Additionally, we evaluate the effectiveness of our in-the-wild pipeline by comparing it with a scenario using ground truth contact information.
The results indicate a modest 2.5\%p gap in mIoU, demonstrating the robustness of our method even in practical, annotation-free scenarios.

\begin{figure}[!t]
\centering
\includegraphics[width=\linewidth]{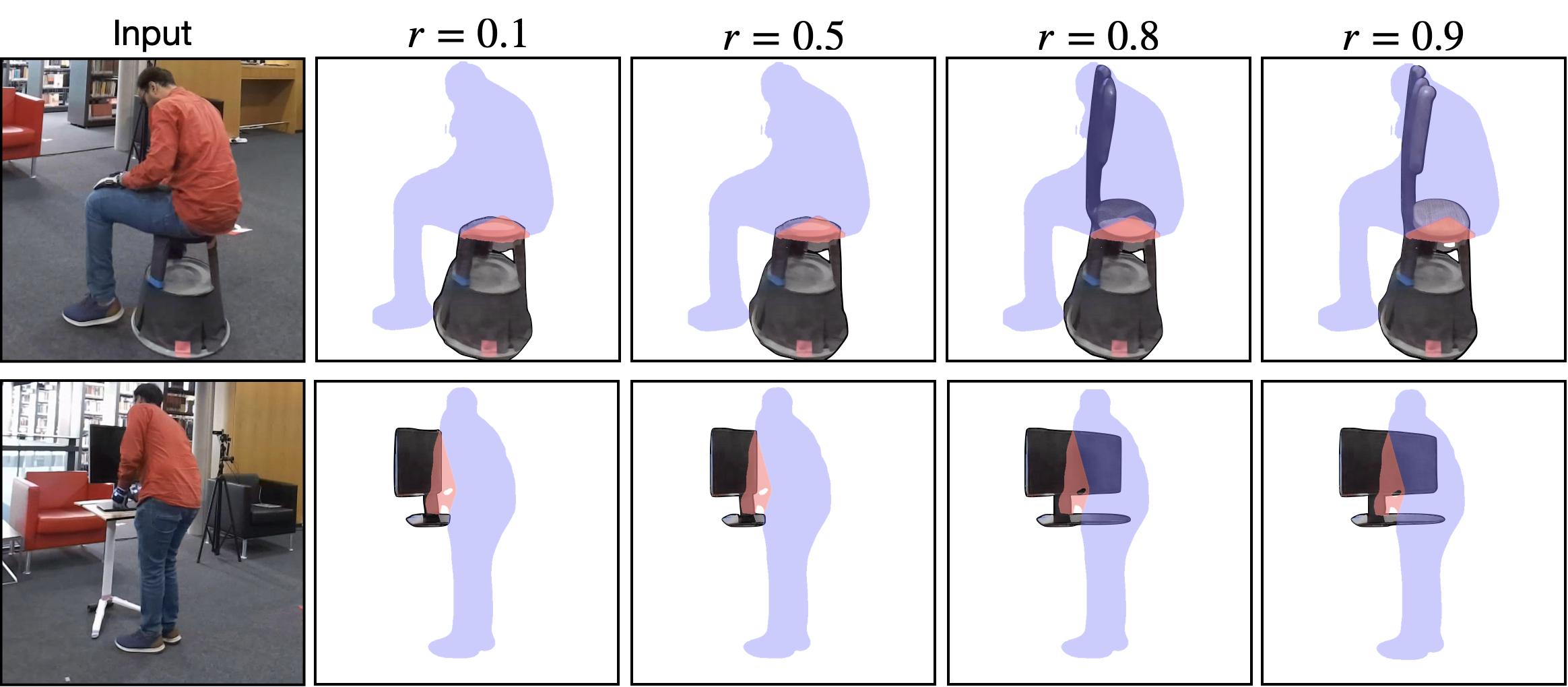}
\caption{
Amodal completion results based on the inpainting strength parameter \( r \). When \( r \) is close to 0, the model focuses mainly on the primary region \textcolor{RedOrange}{\( M_p \)} (\textcolor{RedOrange}{orange}). As \( r \) approaches 1, the model extends its attention to include both \textcolor{RedOrange}{\( M_p \)} (\textcolor{RedOrange}{orange}) and the secondary region \textcolor{Violet}{\( M_s \)} (\textcolor{Violet}{violet}).}
\label{fig:r_qualitative}
\end{figure}

\vspace{-1em}
\paragraph{Effect of Strength Parameter on Amodal Completion}
We evaluate the impact of the strength parameter \( r \) on amodal completion performance through an ablation study using the BEHAVE dataset, as shown in~\cref{fig:r_qualitative} and~\cref{tab:r}. Occlusion cases are divided into two groups based on occlusion ratio: light occlusion (10–40\%) and heavy occlusion (40–70\%). For both groups, varying \( r \) affects the CLIP and mIoU scores, but with opposite tendencies. When the occluded area is small (top row in~\cref{fig:r_qualitative} and left columns in~\cref{tab:r}), a smaller \( r \) yields better mIoU performance. Conversely, when the occluded area is large (bottom row in~\cref{fig:r_qualitative} and middle columns in~\cref{tab:r}), a larger \( r \) tends to produce superior results. This is because a larger \( r \) facilitates inpainting a broader area, including the secondary region $M_s$, while a smaller \( r \) primarily focuses on the primary region $M_p$, leaving insufficient steps to inpaint $M_s$. Given these trends, \( r=0.5 \) generally provides the best overall performance, suggesting it as a balanced value when the occlusion ratio is unknown.

\begin{table}[!t]
\centering
\resizebox{\linewidth}{!}{
\begin{tabular}{l c c c c c c}
\hline
 & \multicolumn{2}{c}{Occ. (10-40\%)} & \multicolumn{2}{c}{Occ. (40-70\%)} & \multicolumn{2}{c}{Total}\\
 \cline{2-3}\cline{4-5}\cline{6-7}
\textbf{Method} & CLIP $\uparrow$ & mIoU $\uparrow$ & CLIP $\uparrow$ & mIoU $\uparrow$ & CLIP $\uparrow$ & mIoU $\uparrow$ \\
\hline
\hline

$r=1.00$ & 26.37 & 72.45\%  & 26.11 & 68.33\%  &  26.27 & 69.98\% \\
$r=0.90$ & \textbf{27.01} & 80.33\%  & \textbf{26.94} & \textbf{73.94\%}  &  \textbf{26.97}  & 76.50\% \\
$r=0.50$ & 27.00 & 84.70\%  & 26.85 & 72.93\%  &  26.91  & \textbf{77.64\%} \\
$r=0.10$ & 26.94 & \textbf{85.44\%}  & 26.82 & 71.54\%  &  26.87  & 77.10\% \\
$r=0.00$ & 26.82 & 84.97\%  & 26.50 & 70.20\%  &  26.63  & 76.11\% \\
\hline
\end{tabular}
}
\caption{Ablation study on mask strength parameter, grouped by occlusion ratio, for amodal completion on the BEHAVE dataset.}
\label{tab:r}
\end{table}

\begin{figure}[t]
\centering
\includegraphics[width=1.00\linewidth]{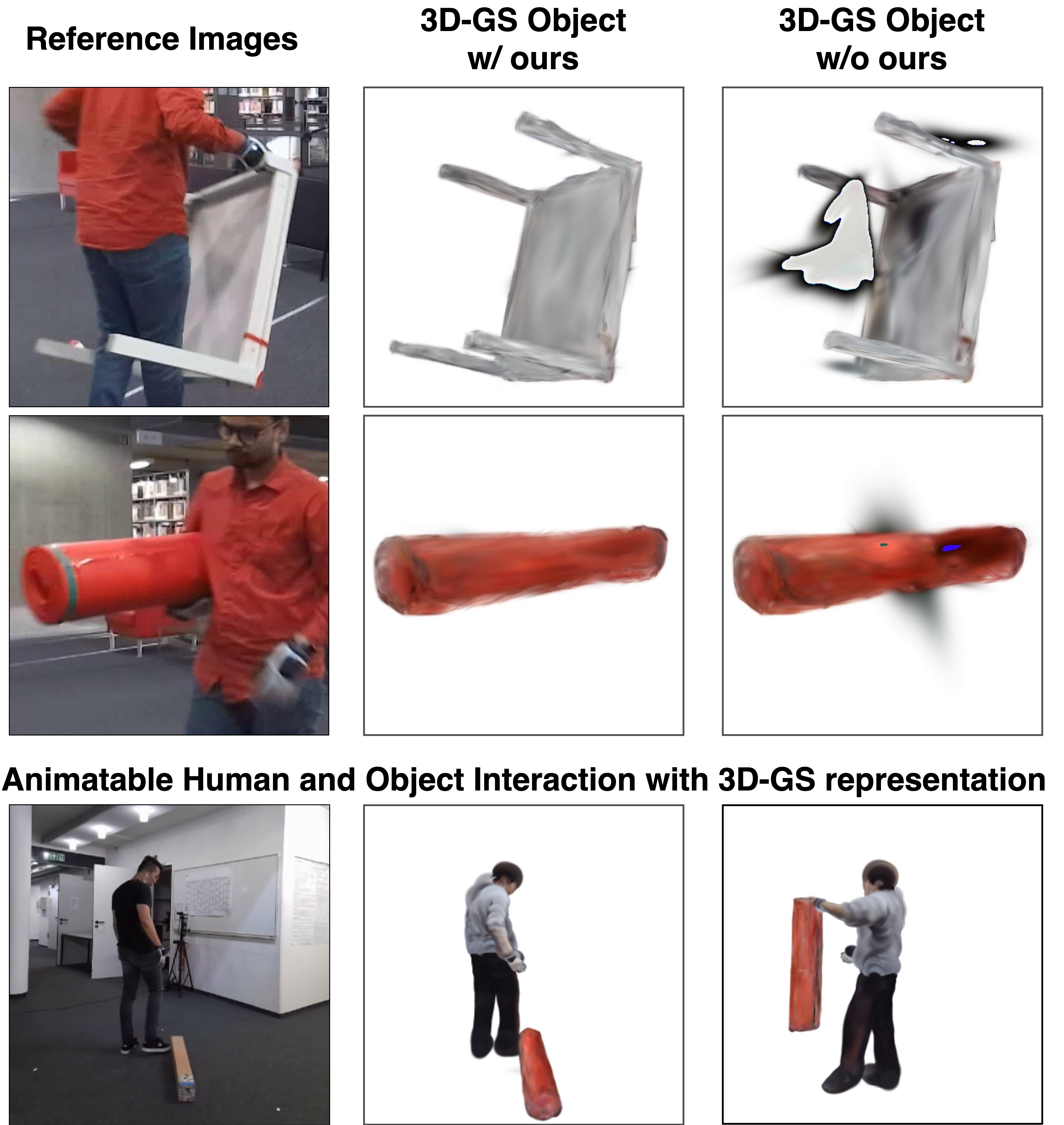}
\caption{Our amodal completion method enhances 3D Gaussian Splatting and extends to joint human-object novel-pose/view synthesis. The last row demonstrates that separately trained 3D-GS human and 3D-GS object can be animated with novel poses extracted from another video.}
\label{fig:gs_obj_avatar}
\end{figure}

\subsection{Applications}
\begin{figure}[t]
\centering
\includegraphics[width=\linewidth]{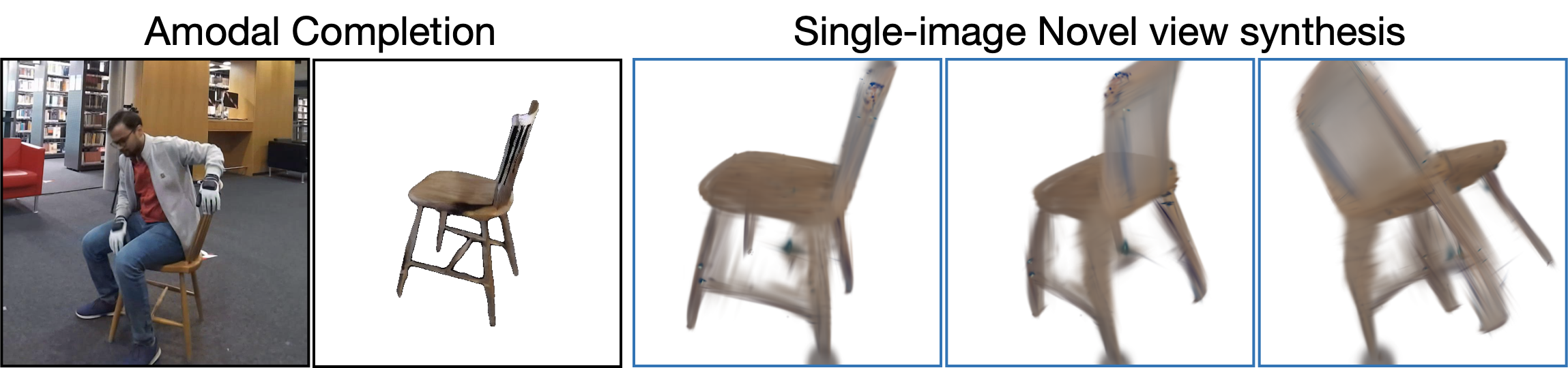}
\caption{Single-view 3D reconstruction results using Triplane~\cite{zou2024triplane}. Our amodal completion acts as a bridge, transforming occluded images into inputs suitable for single-view reconstruction models.}
\label{fig:single_view}
\end{figure}

Our amodal completion method can be extended to enhance various tasks. To demonstrate its utility, we apply 3D reconstruction on objects in the BEHAVE dataset using 3D Gaussian Splatting (3D-GS)~\cite{kerbl20233d} for novel pose synthesis with multi-view setup.
Especially, for human 3D-GS, we follow the method of Guassian Avatar~\cite{hu2024gaussianavatar}.
Due to frequent occlusions from human-object interactions in BEHAVE, training 3D-GS from HOI scenarios is challenging. However, as shown in Figure \ref{fig:gs_obj_avatar}, comparing original images with amodal completed images reveals that our method significantly improves the quality of the trained 3D-GS for object.
In addition, we demonstrate the potential of our amodal completion method for enabling joint human-object novel-pose synthesis and novel-view synthesis, showcasing its ability to effectively handle complex interactions and occlusions, thereby broadening its applicability to more challenging real-world scenarios.
Finally, we validate the versatility of our method on single-view 3D reconstruction, as presented in~\cref{tab:3D_recon} and~\cref{fig:single_view}, using the Triplane~\cite{zou2024triplane}. For implementation details and additional qualitative examples of these applications, please refer to the supplementary material.




\section{Discussion and Limitation}\label{sec:conclusion}

Our work may have limitations in generalizing to scenarios with multiple subjects occluding each other. The dataset employed in our study primarily consists of indoor scenes featuring single human-object interactions, so our method might not generalize well to environments with several humans and objects. Moreover, our approach is designed for single-image processing and is affected by the stochastic nature of diffusion models, leading to a lack of temporal consistency that restricts its application in video tasks requiring frame-to-frame coherence.
Additionally, our model relies heavily on the inpainting capabilities of the diffusion model, which may struggle to reconstruct objects that were not seen during the training of the stable diffusion model.

\begin{table}[t!]
\centering
\resizebox{0.65\linewidth}{!}{
\begin{tabular}{l c}
\hline
\textbf{Method} & CD $\downarrow$ \\
\hline
\hline
SAM~\cite{ravi2024sam} + Triplane~\cite{zou2024triplane} & 0.2303 \\
pix2gestalt~\cite{ozguroglu2024pix2gestalt} + Triplane~\cite{zou2024triplane} & 0.2258 \\
Ours + Triplane~\cite{zou2024triplane} & 0.2155 \\
Ours (GT mask) + Triplane~\cite{zou2024triplane} & 0.2089 \\
\hline
\end{tabular}
}
\caption{3D mesh reconstruction of object with single-view image.}
\label{tab:3D_recon}
\end{table}

\section{Conclusion}
To summarize, we have presented a novel approach to amodal completion that markedly improves the realism and precision of reconstructing occluded object appearances, especially within complex human-object interaction settings. Our method utilizes a multi-regional inpainting strategy that incorporates physical constraints and contact information to delineate regions with different occlusion probabilities, thus enabling focused denoising within the diffusion model. By effectively addressing both structural and visual components, our approach moves artificial perception closer to a more intuitive, human-like interpretation of occluded scenes. Our experimental results confirm that the proposed method surpasses existing techniques, demonstrating its robustness and efficacy in HOI scenarios even in the absence of ground-truth annotations.

While our work focuses on single images, future extensions could address its current limitations, such as generalizing to more complicated scenarios involving multiple humans and objects or incorporating temporal consistency to handle video data. Expanding the approach to account for dynamic sequences would enable realistic and coherent reconstructions across frames, further broadening its applicability to challenging real-world settings. This direction holds promise for advancing 3D HOI reconstruction and enriching applications in AR/VR and robotics.

\clearpage
{
    \small
    \bibliographystyle{ieeenat_fullname}
    \bibliography{main}

\begin{thebibliography}{45}
\providecommand{\natexlab}[1]{#1}
\providecommand{\url}[1]{\texttt{#1}}
\expandafter\ifx\csname urlstyle\endcsname\relax
  \providecommand{\doi}[1]{doi: #1}\else
  \providecommand{\doi}{doi: \begingroup \urlstyle{rm}\Url}\fi

\bibitem[Baradel et~al.(2024)Baradel, Armando, Galaaoui, Br{\'e}gier, Weinzaepfel, Rogez, and Lucas]{baradel2024multi}
Fabien Baradel, Matthieu Armando, Salma Galaaoui, Romain Br{\'e}gier, Philippe Weinzaepfel, Gr{\'e}gory Rogez, and Thomas Lucas.
\newblock Multi-hmr: Multi-person whole-body human mesh recovery in a single shot.
\newblock In \emph{European Conference on Computer Vision}, pages 202--218. Springer, 2024.

\bibitem[Bhatnagar et~al.(2022)Bhatnagar, Xie, Petrov, Sminchisescu, Theobalt, and Pons-Moll]{bhatnagar2022behave}
Bharat~Lal Bhatnagar, Xianghui Xie, Ilya~A Petrov, Cristian Sminchisescu, Christian Theobalt, and Gerard Pons-Moll.
\newblock Behave: Dataset and method for tracking human object interactions.
\newblock In \emph{Proceedings of the IEEE/CVF Conference on Computer Vision and Pattern Recognition}, pages 15935--15946, 2022.

\bibitem[Chen et~al.(2016)Chen, M{\"u}ller, and Conci]{chen2016amodal}
Siyi Chen, Hermann~J M{\"u}ller, and Markus Conci.
\newblock Amodal completion in visual working memory.
\newblock \emph{Journal of Experimental Psychology: Human Perception and Performance}, 42\penalty0 (9):\penalty0 1344, 2016.

\bibitem[Chen et~al.(2023)Chen, Dwivedi, Black, and Tzionas]{chen2023detecting}
Yixin Chen, Sai~Kumar Dwivedi, Michael~J Black, and Dimitrios Tzionas.
\newblock Detecting human-object contact in images.
\newblock In \emph{Proceedings of the IEEE/CVF Conference on Computer Vision and Pattern Recognition}, pages 17100--17110, 2023.

\bibitem[Emmanouil and Ro(2014)]{emmanouil2014amodal}
Tatiana~Aloi Emmanouil and Tony Ro.
\newblock Amodal completion of unconsciously presented objects.
\newblock \emph{Psychonomic Bulletin \& Review}, 21:\penalty0 1188--1194, 2014.

\bibitem[Gao et~al.(2023)Gao, Qian, Wang, Xiao, He, Zhang, and Fu]{gao2023coarse}
Jianxiong Gao, Xuelin Qian, Yikai Wang, Tianjun Xiao, Tong He, Zheng Zhang, and Yanwei Fu.
\newblock Coarse-to-fine amodal segmentation with shape prior.
\newblock In \emph{Proceedings of the IEEE/CVF International Conference on Computer Vision}, pages 1262--1271, 2023.

\bibitem[Hu et~al.(2024)Hu, Zhang, Zhang, Zhou, Liu, Zhang, and Nie]{hu2024gaussianavatar}
Liangxiao Hu, Hongwen Zhang, Yuxiang Zhang, Boyao Zhou, Boning Liu, Shengping Zhang, and Liqiang Nie.
\newblock Gaussianavatar: Towards realistic human avatar modeling from a single video via animatable 3d gaussians.
\newblock In \emph{Proceedings of the IEEE/CVF Conference on Computer Vision and Pattern Recognition}, pages 634--644, 2024.

\bibitem[Huang et~al.(2022)Huang, Taheri, Black, and Tzionas]{huang2022intercap}
Yinghao Huang, Omid Taheri, Michael~J Black, and Dimitrios Tzionas.
\newblock Intercap: Joint markerless 3d tracking of humans and objects in interaction.
\newblock In \emph{DAGM German Conference on Pattern Recognition}, pages 281--299. Springer, 2022.

\bibitem[Jayaram and Fleyeh(2016)]{jayaram2016convex}
MA Jayaram and Hasan Fleyeh.
\newblock Convex hulls in image processing: a scoping review.
\newblock \emph{American Journal of Intelligent Systems}, 6\penalty0 (2):\penalty0 48--58, 2016.

\bibitem[Karavelas et~al.(2013)Karavelas, Seidel, and Tzanaki]{karavelas2013convex}
Menelaos~I Karavelas, Raimund Seidel, and Eleni Tzanaki.
\newblock Convex hulls of spheres and convex hulls of disjoint convex polytopes.
\newblock \emph{Computational Geometry}, 46\penalty0 (6):\penalty0 615--630, 2013.

\bibitem[Ke et~al.(2021)Ke, Tai, and Tang]{ke2021deep}
Lei Ke, Yu-Wing Tai, and Chi-Keung Tang.
\newblock Deep occlusion-aware instance segmentation with overlapping bilayers.
\newblock In \emph{Proceedings of the IEEE/CVF conference on computer vision and pattern recognition}, pages 4019--4028, 2021.

\bibitem[Kerbl et~al.(2023)Kerbl, Kopanas, Leimkuehler, and Drettakis]{kerbl20233d}
Bernhard Kerbl, Georgios Kopanas, Thomas Leimkuehler, and George Drettakis.
\newblock 3d gaussian splatting for real-time radiance field rendering.
\newblock \emph{ACM Transactions on Graphics (TOG)}, 42\penalty0 (4):\penalty0 1--14, 2023.

\bibitem[Kim et~al.(2024)Kim, Han, Kwon, and Joo]{coma}
Hyeonwoo Kim, Sookwan Han, Patrick Kwon, and Hanbyul Joo.
\newblock Beyond the contact: Discovering comprehensive affordance for 3d objects from pre-trained 2d diffusion models, 2024.

\bibitem[Lee and Park(2022)]{lee2022instaorder}
Hyunmin Lee and Jaesik Park.
\newblock {Instance-wise Occlusion and Depth Orders in Natural Scenes}.
\newblock In \emph{Proceedings of the {IEEE} Conference on Computer Vision and Pattern Recognition}, 2022.

\bibitem[Li and Dai(2024)]{li2024genzi}
Lei Li and Angela Dai.
\newblock Genzi: Zero-shot 3d human-scene interaction generation.
\newblock In \emph{Proceedings of the IEEE/CVF Conference on Computer Vision and Pattern Recognition}, pages 20465--20474, 2024.

\bibitem[Ling et~al.(2020)Ling, Acuna, Kreis, Kim, and Fidler]{ling2020variational}
Huan Ling, David Acuna, Karsten Kreis, Seung~Wook Kim, and Sanja Fidler.
\newblock Variational amodal object completion.
\newblock \emph{Advances in Neural Information Processing Systems}, 33:\penalty0 16246--16257, 2020.

\bibitem[Liu et~al.(2023)Liu, Li, Wu, and Lee]{liu2023llava}
Haotian Liu, Chunyuan Li, Qingyang Wu, and Yong~Jae Lee.
\newblock Visual instruction tuning.
\newblock In \emph{NeurIPS}, 2023.

\bibitem[Loper et~al.(2023)Loper, Mahmood, Romero, Pons-Moll, and Black]{loper2023smpl}
Matthew Loper, Naureen Mahmood, Javier Romero, Gerard Pons-Moll, and Michael~J Black.
\newblock Smpl: A skinned multi-person linear model.
\newblock In \emph{Seminal Graphics Papers: Pushing the Boundaries, Volume 2}, pages 851--866. 2023.

\bibitem[Moon et~al.(2022)Moon, Choi, and Lee]{moon2022accurate}
Gyeongsik Moon, Hongsuk Choi, and Kyoung~Mu Lee.
\newblock Accurate 3d hand pose estimation for whole-body 3d human mesh estimation.
\newblock In \emph{Proceedings of the IEEE/CVF Conference on Computer Vision and Pattern Recognition}, pages 2308--2317, 2022.

\bibitem[Nam et~al.(2024)Nam, Jung, Moon, and Lee]{nam2024joint}
Hyeongjin Nam, Daniel~Sungho Jung, Gyeongsik Moon, and Kyoung~Mu Lee.
\newblock Joint reconstruction of 3d human and object via contact-based refinement transformer.
\newblock In \emph{Proceedings of the IEEE/CVF Conference on Computer Vision and Pattern Recognition}, pages 10218--10227, 2024.

\bibitem[OpenAI(2024)]{openai2024chatgpt4}
OpenAI.
\newblock Chatgpt-4.
\newblock \url{https://openai.com/}, 2024.
\newblock Large language model.

\bibitem[Ozguroglu et~al.(2024)Ozguroglu, Liu, Sur{\'\i}s, Chen, Dave, Tokmakov, and Vondrick]{ozguroglu2024pix2gestalt}
Ege Ozguroglu, Ruoshi Liu, D{\'\i}dac Sur{\'\i}s, Dian Chen, Achal Dave, Pavel Tokmakov, and Carl Vondrick.
\newblock pix2gestalt: Amodal segmentation by synthesizing wholes.
\newblock In \emph{2024 IEEE/CVF Conference on Computer Vision and Pattern Recognition (CVPR)}, pages 3931--3940. IEEE Computer Society, 2024.

\bibitem[Radford et~al.(2021)Radford, Kim, Hallacy, Ramesh, Goh, Agarwal, Sastry, Askell, Mishkin, Clark, et~al.]{radford2021learning}
Alec Radford, Jong~Wook Kim, Chris Hallacy, Aditya Ramesh, Gabriel Goh, Sandhini Agarwal, Girish Sastry, Amanda Askell, Pamela Mishkin, Jack Clark, et~al.
\newblock Learning transferable visual models from natural language supervision.
\newblock In \emph{International conference on machine learning}, pages 8748--8763. PMLR, 2021.

\bibitem[Ravi et~al.(2024)Ravi, Gabeur, Hu, Hu, Ryali, Ma, Khedr, R{\"a}dle, Rolland, Gustafson, et~al.]{ravi2024sam}
Nikhila Ravi, Valentin Gabeur, Yuan-Ting Hu, Ronghang Hu, Chaitanya Ryali, Tengyu Ma, Haitham Khedr, Roman R{\"a}dle, Chloe Rolland, Laura Gustafson, et~al.
\newblock Sam 2: Segment anything in images and videos.
\newblock \emph{arXiv preprint arXiv:2408.00714}, 2024.

\bibitem[Rombach et~al.(2022)Rombach, Blattmann, Lorenz, Esser, and Ommer]{rombach2022high}
Robin Rombach, Andreas Blattmann, Dominik Lorenz, Patrick Esser, and Bj{\"o}rn Ommer.
\newblock High-resolution image synthesis with latent diffusion models.
\newblock In \emph{Proceedings of the IEEE/CVF conference on computer vision and pattern recognition}, pages 10684--10695, 2022.

\bibitem[Rosin(2000)]{rosin2000shape}
Paul~L Rosin.
\newblock Shape partitioning by convexity.
\newblock \emph{IEEE Transactions on Systems, Man, and Cybernetics-Part A: Systems and Humans}, 30\penalty0 (2):\penalty0 202--210, 2000.

\bibitem[Serra(1983)]{serra1983image}
J Serra.
\newblock Image analysis and mathematical morphology, 1983.

\bibitem[Sirakov and Mlsna(2004)]{sirakov2004search}
Nikolay~M Sirakov and Phillip~A Mlsna.
\newblock Search space partitioning using convex hull and concavity features for fast medical image retrieval.
\newblock In \emph{2004 2nd IEEE International Symposium on Biomedical Imaging: Nano to Macro (IEEE Cat No. 04EX821)}, pages 796--799. IEEE, 2004.

\bibitem[Song et~al.(2020)Song, Meng, and Ermon]{song2020denoising}
Jiaming Song, Chenlin Meng, and Stefano Ermon.
\newblock Denoising diffusion implicit models.
\newblock \emph{arXiv preprint arXiv:2010.02502}, 2020.

\bibitem[Sun et~al.(2022)Sun, Kortylewski, and Yuille]{sun2022amodal}
Yihong Sun, Adam Kortylewski, and Alan Yuille.
\newblock Amodal segmentation through out-of-task and out-of-distribution generalization with a bayesian model.
\newblock In \emph{Proceedings of the IEEE/CVF Conference on Computer Vision and Pattern Recognition}, pages 1215--1224, 2022.

\bibitem[Suvorov et~al.(2022)Suvorov, Logacheva, Mashikhin, Remizova, Ashukha, Silvestrov, Kong, Goka, Park, and Lempitsky]{suvorov2022resolution}
Roman Suvorov, Elizaveta Logacheva, Anton Mashikhin, Anastasia Remizova, Arsenii Ashukha, Aleksei Silvestrov, Naejin Kong, Harshith Goka, Kiwoong Park, and Victor Lempitsky.
\newblock Resolution-robust large mask inpainting with fourier convolutions.
\newblock In \emph{Proceedings of the IEEE/CVF winter conference on applications of computer vision}, pages 2149--2159, 2022.

\bibitem[Tripathi et~al.(2023)Tripathi, Chatterjee, Passy, Yi, Tzionas, and Black]{tripathi2023deco}
Shashank Tripathi, Agniv Chatterjee, Jean-Claude Passy, Hongwei Yi, Dimitrios Tzionas, and Michael~J. Black.
\newblock {DECO}: Dense estimation of {3D} human-scene contact in the wild.
\newblock In \emph{Proceedings of the IEEE/CVF International Conference on Computer Vision (ICCV)}, pages 8001--8013, 2023.

\bibitem[Wang et~al.(2014)Wang, Emmerich, Li, Tang, B{\"a}ck, and Yao]{wang2014convex}
Pu Wang, Michael Emmerich, Rui Li, Ke Tang, Thomas B{\"a}ck, and Xin Yao.
\newblock Convex hull-based multiobjective genetic programming for maximizing receiver operating characteristic performance.
\newblock \emph{IEEE Transactions on Evolutionary Computation}, 19\penalty0 (2):\penalty0 188--200, 2014.

\bibitem[Xie et~al.(2022)Xie, Bhatnagar, and Pons-Moll]{xie2022chore}
Xianghui Xie, Bharat~Lal Bhatnagar, and Gerard Pons-Moll.
\newblock Chore: Contact, human and object reconstruction from a single rgb image.
\newblock In \emph{European Conference on Computer Vision ({ECCV})}. {Springer}, 2022.

\bibitem[Xie et~al.(2023)Xie, Bhatnagar, and Pons-Moll]{xie2023vistracker}
Xianghui Xie, Bharat~Lal Bhatnagar, and Gerard Pons-Moll.
\newblock Visibility aware human-object interaction tracking from single rgb camera.
\newblock In \emph{IEEE Conference on Computer Vision and Pattern Recognition (CVPR)}, 2023.

\bibitem[Xie et~al.(2024)Xie, Bhatnagar, Lenssen, and Pons-Moll]{xie2023template_free}
Xianghui Xie, Bharat~Lal Bhatnagar, Jan~Eric Lenssen, and Gerard Pons-Moll.
\newblock Template free reconstruction of human-object interaction with procedural interaction generation.
\newblock In \emph{IEEE Conference on Computer Vision and Pattern Recognition (CVPR)}, 2024.

\bibitem[Xu et~al.(2024)Xu, Zhang, and Shi]{xu2024amodal}
Katherine Xu, Lingzhi Zhang, and Jianbo Shi.
\newblock Amodal completion via progressive mixed context diffusion.
\newblock In \emph{Proceedings of the IEEE/CVF Conference on Computer Vision and Pattern Recognition}, pages 9099--9109, 2024.

\bibitem[Yang et~al.(2013)Yang, Zhang, and Lu]{yang2013graph}
Chuan Yang, Lihe Zhang, and Huchuan Lu.
\newblock Graph-regularized saliency detection with convex-hull-based center prior.
\newblock \emph{IEEE Signal Processing Letters}, 20\penalty0 (7):\penalty0 637--640, 2013.

\bibitem[Yang et~al.(2024)Yang, Zhai, Luo, Cao, and Zha]{yang2024lemon}
Yuhang Yang, Wei Zhai, Hongchen Luo, Yang Cao, and Zheng-Jun Zha.
\newblock Lemon: Learning 3d human-object interaction relation from 2d images.
\newblock In \emph{Proceedings of the IEEE/CVF Conference on Computer Vision and Pattern Recognition}, pages 16284--16295, 2024.

\bibitem[Yildirim et~al.(2023)Yildirim, Baday, Erdem, Erdem, and Dundar]{yildirim2023inst}
Ahmet~Burak Yildirim, Vedat Baday, Erkut Erdem, Aykut Erdem, and Aysegul Dundar.
\newblock Inst-inpaint: Instructing to remove objects with diffusion models.
\newblock \emph{arXiv preprint arXiv:2304.03246}, 2023.

\bibitem[Zhan et~al.(2024)Zhan, Zheng, Xie, and Zisserman]{zhan2024amodal}
Guanqi Zhan, Chuanxia Zheng, Weidi Xie, and Andrew Zisserman.
\newblock Amodal ground truth and completion in the wild.
\newblock In \emph{Proceedings of the IEEE/CVF Conference on Computer Vision and Pattern Recognition}, pages 28003--28013, 2024.

\bibitem[Zhan et~al.(2020)Zhan, Pan, Dai, Liu, Lin, and Loy]{zhan2020self}
Xiaohang Zhan, Xingang Pan, Bo Dai, Ziwei Liu, Dahua Lin, and Chen~Change Loy.
\newblock Self-supervised scene de-occlusion.
\newblock In \emph{Proceedings of the IEEE conference on computer vision and pattern recognition (CVPR)}, 2020.

\bibitem[Zhang et~al.(2024)Zhang, Liu, Zhang, Wang, Liu, Lin, and Liu]{zhang2024amodal}
Bowen Zhang, Qing Liu, Jianming Zhang, Yilin Wang, Liyang Liu, Zhe Lin, and Yifan Liu.
\newblock Amodal scene analysis via holistic occlusion relation inference and generative mask completion.
\newblock In \emph{Proceedings of the AAAI Conference on Artificial Intelligence}, pages 6997--7005, 2024.

\bibitem[Zhu et~al.(2024)Zhu, Li, Tang, Zhao, Zhou, and Lu]{zhu2024dpmesh}
Yixuan Zhu, Ao Li, Yansong Tang, Wenliang Zhao, Jie Zhou, and Jiwen Lu.
\newblock Dpmesh: Exploiting diffusion prior for occluded human mesh recovery.
\newblock In \emph{Proceedings of the IEEE/CVF Conference on Computer Vision and Pattern Recognition}, pages 1101--1110, 2024.

\bibitem[Zou et~al.(2024)Zou, Yu, Guo, Li, Liang, Cao, and Zhang]{zou2024triplane}
Zi-Xin Zou, Zhipeng Yu, Yuan-Chen Guo, Yangguang Li, Ding Liang, Yan-Pei Cao, and Song-Hai Zhang.
\newblock Triplane meets gaussian splatting: Fast and generalizable single-view 3d reconstruction with transformers.
\newblock In \emph{Proceedings of the IEEE/CVF Conference on Computer Vision and Pattern Recognition}, pages 10324--10335, 2024.

\end{thebibliography}
}

\clearpage
\setcounter{page}{1}
\maketitlesupplementary

\appendix

\begin{table}[t]
\centering
\resizebox{0.7\linewidth}{!}{
\begin{tabular}{l c c}
\hline
\textbf{Region} & BEHAVE & InterCap \\
\hline
\hline
Primary Region & 48.07 \% & 35.05 \%\\ 
Secondary Region & \ \ 6.74 \% & \ \ 2.83 \%\\

\hline
\end{tabular}
}
\caption{Average percentage of occluded pixels in the primary and secondary regions for the BEHAVE and InterCap datasets.}
\vspace{-1.em}
\label{tab:region_ratio}
\end{table}

\section{Additional Details}
We use the default parameters for all baselines and pre-trained models unless specified otherwise.

\subsection{Occluded Pixel Ratios in Multi-Regions}
\Cref{tab:region_ratio} presents the percentage of occluded pixels within the primary and secondary regions. The percentage is computed based on the 2D area as follows:

\begin{equation}
\frac{|M^\text{full}_\text{obj} \cap M_\text{region}|}{|M_\text{region}|},
\label{eq:occlusion_percentage}
\end{equation}
where \( M^\text{full}_\text{obj} \) denotes the projection of the fully rendered 3D object in image space, \( M_\text{region} \) corresponds to either the primary or secondary region, and \( || \) represents the area of the mask, calculated by summing the binary mask values along the width and height axes.

In the BEHAVE dataset, the primary region effectively covers the inpainting area, with 48.07\% of the primary region containing occluded parts. In contrast, the secondary region accounts for only 6.74\%, emphasizing the need for careful handling of the secondary region.

\subsection{Data Selection}
For both the BEHAVE~\cite{bhatnagar2022behave} and InterCap~\cite{huang2022intercap} datasets, we filter out images where the object occlusion is either less than 10\% or greater than 70\%, as these extremes provide limited value for evaluating occlusion handling. Additionally, we exclude frames where the visible area of the object is less than 5\% of the human mask, ensuring sufficient detail for reliable analysis. These criteria maintain a balanced and robust dataset for evaluating our methods.

\subsection{Implementation Details}

\paragraph{Dataloader}
For the BEHAVE dataset, we utilized the dataloader provided by the HDM~\cite{xie2023template_free} GitHub repository (\url{https://github.com/xiexh20/HDM}). Based on this BEHAVE dataloader, we preprocess the InterCap~\cite{huang2022intercap} dataset to follow the same structure as the BEHAVE dataset, ensuring compatibility with minimal modifications to the original dataloader from HDM.

\vspace{-0.5em}
\paragraph{Baselines}
\begin{itemize}
    \item \textbf{Pix2Gestalt}~\cite{ozguroglu2024pix2gestalt}: We borrow the code and pre-trained model from \url{https://github.com/cvlab-columbia/pix2gestalt} and adapt it to be compatible with our dataloader implementation. Pix2Gestalt requires only the segmented image for amodal completion.
    \item \textbf{Xu et al.}~\cite{xu2024amodal}: To ensure a fair zero-shot comparison, we made several modifications to the code borrowed from \url{https://github.com/k8xu/amodal}. Since the original code was designed for 83 specific object classes, we replaced its InstaOrder~\cite{lee2022instaorder} module with ground-truth depth ordering, supplied explicit occluder/occludee segmentation masks, and constrained its multi-iteration scheme to a single pass.
    \item \textbf{LaMa}~\cite{suvorov2022resolution}: We utilize the code from \url{https://github.com/enesmsahin/simple-lama-inpainting}. LaMa requires the original image and the occluder mask to perform inpainting.
    \item \textbf{Inst-Inpaint}~\cite{yildirim2023inst}: We borrow the code and pre-trained model from \url{https://github.com/abyildirim/inst-inpaint}. Inst-Inpaint requires the original image and a text prompt specifying the object to remove. For example, "remove the person in the center."
    \item \textbf{Naive Outpainting}~\cite{rombach2022high}: We employee the SD-inpaint model from \url{https://github.com/huggingface/diffusers}, which requires a segmented image and an inpaint mask. Here, the inpaint mask is defined as the remaining area outside the segmented image.
\end{itemize}

\vspace{-1.5em}
\paragraph{Application}
To demonstrate that our amodal completion method enhances downstream tasks like 3D reconstruction, we explored human-object interaction reconstruction, consisting of animatable human avatar creation and 3D object reconstruction.

We conducted both tasks on the BEHAVE dataset, which provides sequences with four synchronized views, ground truth SMPLH poses, and object poses for each timestamp. For simplicity, we used only a single view in both tasks.

For animatable human avatar creation, we followed the approach of GaussianAvatar~\cite{hu2024gaussianavatar}. Using single-view data and the provided ground truth SMPLH poses as input, we trained the human avatar model.

For 3D object reconstruction, we applied 3D Gaussian Splatting (3DGS)~\cite{kerbl20233d} to reconstruct moving objects from a single view. Since our setting involves a fixed camera with moving objects—unlike the original 3DGS setup with a static scene and moving camera—we adapted 3DGS by treating the object's pose as the inverse of the camera's pose.

Comparing results using the original occluded images versus the amodally completed images in both tasks demonstrated the effectiveness of our amodal completion method in enhancing 3D reconstruction as shown in~\Cref{suppl:3D_recon}.

\subsection{Pseudo Code for Multi-Regional Inpainting}
We present the pseudo code for Multi-Regional Inpainting in~\Cref{algorithm:multi-regional}, which outlines the key steps for handling multiple regions with varying occlusion levels. This approach ensures accurate and context-aware reconstruction by prioritizing regions based on occlusion characteristics. For full technical details and reproducibility, the complete implementation is included as an attached file.

\begin{figure}[h]
\centering
\vspace{-1.5em}
\resizebox{!}{0.75\linewidth}{
    \begin{minipage}{0.56\textwidth} 
        \begin{algorithm}[H]
        \caption{Multi-regional Inpainting}
        \begin{algorithmic}[1]
        \Procedure{Multi-regional Inpaint}{$p, I_\text{in}, M_p, M_s, r, T, S$}
            \State \textbf{Input:} $\mathcal{P}$ (text prompt), $I_\text{in}$ (segmented input image),\\ \quad\ \ $M_p$ (primary mask), $M_s$ (secondary mask), $r$ (strength),\\ \quad\ \ $T$ (maximum timestep), $S$ (scheduler)
            \State \textbf{Output:} Generated inpainted image $I_{\text{out}}$
            
            \State \textbf{Step 1: Prepare Latents}
            \State Initialize latent variable $\ell$ using $I_\text{in}$ and random noise $\eta$
            \State Generate masked latent $\ell_{M_p}$ using $M_p$
            \State Generate masked latent $\ell_{M_p \cup M_s}$ using $M_p$ and $M_s$
            \State Set $T' = \text{int}(T \times r)$ as the maximum timestep for $M_s$
            \State Calculate timesteps $\mathcal{T}$ based on $T$ and $r$
    
            \State \textbf{Step 2: Denoising Process}
            \For{each $t \in \mathcal{T}$}
                \State $\ell_{\text{input}} = \ell$
                \State \textbf{Step 2.1: Scale Latent Model Input}
                \State Scale $\ell_{\text{input}}$ using scheduler $S$ with current timestep $t$
                
                \State \textbf{Step 2.2: Concatenate Inputs for UNet}
                \If{$t > T'$}
                    \State $\ell_{\text{input}} = \text{concat}(\ell_{\text{input}}, M_p, \ell_{M_p})$
                \Else
                    \State $\ell_{\text{input}} = \text{concat}(\ell_{\text{input}}, M_p \cup M_s,  \ell_{M_p \cup M_s})$
                \EndIf
    
                \State \textbf{Step 2.3: Predict Noise Residual}
                \State $\eta' = \text{UNet}(\ell_{\text{input}}, t, \mathcal{P})$
                \State \textbf{Step 2.4: Modify Latent Variable}
                \State Update $\ell$ using guided noise prediction $\eta'$ and scheduler $S$
            \EndFor
            \State \textbf{Step 3: Decode and Post-process}
            \State Decode $\ell$ to generate final image $I_{\text{out}}$
            \State \Return $I_{\text{out}}$
        \EndProcedure
        \end{algorithmic}
        \label{algorithm:multi-regional}
        \end{algorithm}
    \end{minipage}
}
\vspace{-1.5em}
\end{figure}

\section{Additional Analysis on Amodal Completion}\label{sec:Analysis}
\subsection{Human Amodal Completion}\label{sec:human_completion}
While our method is applicable to both human and object amodal completion, we introduce a refined approach specifically for human completion. Leveraging recent advancements in human mesh recovery techniques such as~\cite{zhu2024dpmesh, nam2024joint}, we can accurately delineate occluded regions of human. For human amodal completion, these occluded areas are localized by computing the intersection between the SMPL~\cite{loper2023smpl} body model's projection and the segmentation mask of the interacting object. This targeted approach enables efficient extraction of primary occluded regions, formalized as follows:
\begin{flalign}
    I_\text{out} &= F_{T \rightarrow 0}\left(I^\text{human}_\text{in}, M_\text{smpl} \cap M_\text{obj}, \mathcal{P}\right),
\label{eq:object_second}
\end{flalign}
where \( I^\text{human}_\text{in} \) represents the segmented image of the visible human parts, \( M_\text{smpl} \) is the SMPL body model projection, and \( M_\text{obj} \) denotes the visible object segmentation. This formulation enables precise identification of occluded human regions, allowing for focused and efficient inpainting within the primary occlusion areas.

\begin{figure}[t]
\centering
\begin{center}
    \centering
    \captionsetup{type=figure}
    \vspace{-0.5em}
    \includegraphics[width=0.9\linewidth]{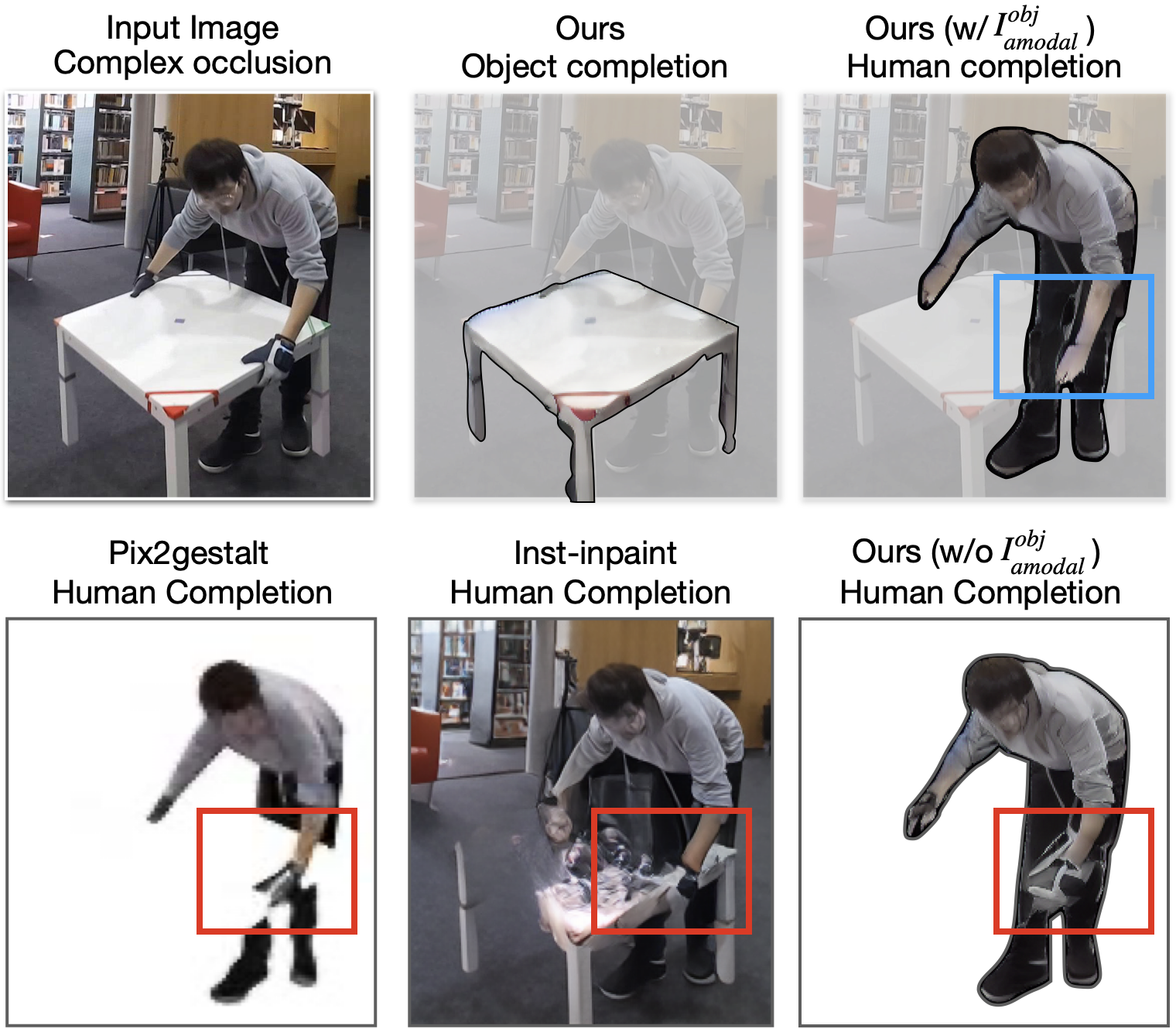}
    \captionof{figure}{Mutual occlusion frequently occurs during HOI due to the dynamic nature of interactions. Baseline models often fail to produce plausible results, as highlighted in the \textcolor{red}{red} box. In contrast, our method generates more coherent results by progressively complete the object and human as shown in the upper row.}
    \label{fig:complex_completion}
\end{center}%
\vspace{-2em}
\end{figure}
\paragraph{Complex Occlusion Scenarios} 
Despite recent advancements, the dynamic nature of human-object interactions often introduces complex occlusions that challenge the quality of amodal completion results. For instance, in~\cref{fig:complex_completion}, when a person interacts with a table, the person's hand and arm occlude parts of the table, while the table simultaneously occludes parts of the person’s legs. Such interactions complicate the accurate reconstruction of occluded human regions, even with topological priors, underscoring the challenges inherent in Human-Object Interaction (HOI) scenarios.
Our observations indicate that repainting the entire region of intersection between the completed object and SMPL projection, rather than inpainting only the occluded areas, frequently yields more coherent and visually plausible results. This approach is captured in the formulation below:
\begin{flalign}
    I_\text{out} &= F_{T \rightarrow 0}(I^\text{human}_\text{in}, M_\text{smpl}\cap\text{Seg}(I^\text{obj}_\text{amodal}), \mathcal{P}),
\label{eq:object_second}
\end{flalign}
where \( I^\text{obj}_\text{amodal} \) represents the amodal completion image of the object, and Seg\( (\cdot)\) represents a segmentation model. In our work, we utilized the Segment Anything Model (SAM)~\cite{ravi2024sam} as the segmentation model. This formulation enables more coherent inpainting by incorporating both the SMPL projection and object segmentation within the amodal completion framework.

\begin{figure}[t!]
    \centering
    \includegraphics[width=\linewidth]{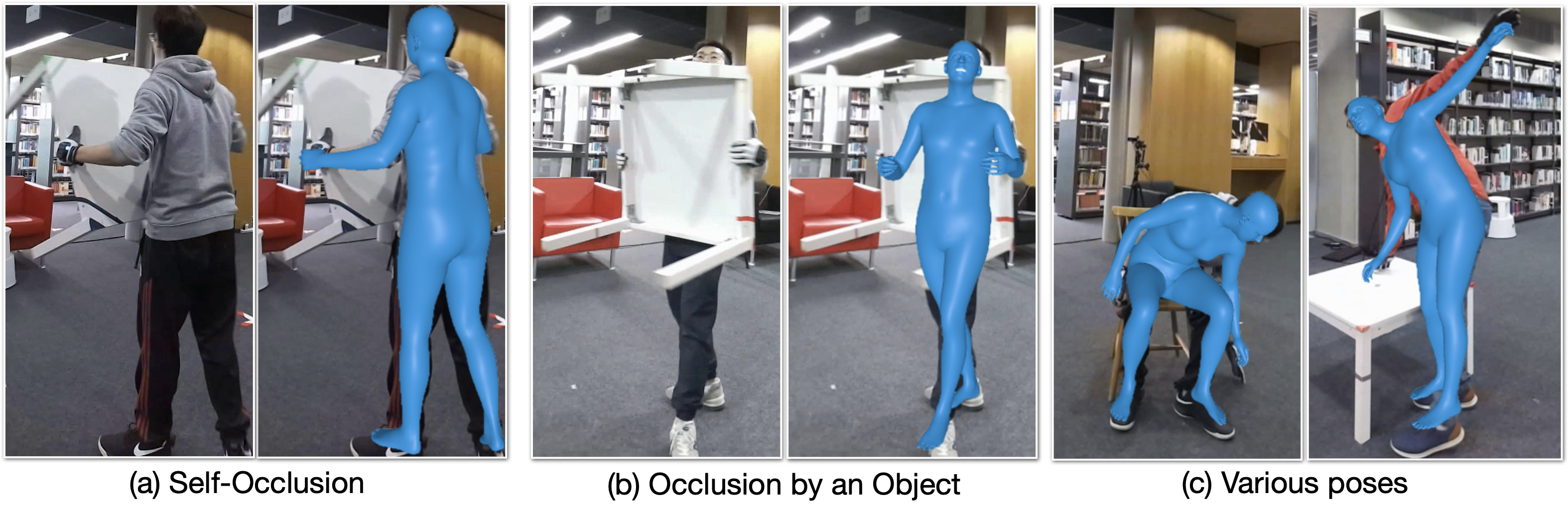}
    \captionsetup{font=scriptsize}
    \vspace{-2.0em}
    \caption{SMPL overlay images obtained by Multi-HMR ~\cite{baradel2024multi} on the BEHAVE.}
    \vspace{-0.8em}
    \captionsetup{font=normalsize}
    \label{fig:multihmr_overlay}
\end{figure}

\begin{table}[t]
\centering
\begin{minipage}[t]{0.645\linewidth}
\centering
    \resizebox{1.0\linewidth}{!}{
    \begin{tabular}{l | c || c c | c c}
        \toprule
        Contact & \multicolumn{1}{c||}{SMPL} & \multicolumn{2}{c|}{Obj. Amodal} & \multicolumn{2}{c}{Human Amodal} \\
        Methods & MPJPE & CLIP $\uparrow$ & mIoU $\uparrow$ & CLIP $\uparrow$ & mIoU $\uparrow$\\
        \hline\hline
        Hand4Whole~\cite{moon2022accurate} & $84.1$ mm & $26.59$ & $74.54\%$ & $27.18$ & $91.35\%$ \\
        DPMesh~\cite{zhu2024dpmesh} & $72.8$ mm & $26.73$ & $76.24\%$ & $27.20$ & $95.23\%$\\
        Multi-HMR~\cite{baradel2024multi} & $68.9$ mm & $26.91$ & $77.64\%$ & $27.21$ & $96.79\%$ \\
        \textbf{GT-contact} & $-$ & $27.07$ & $80.15\%$ & $27.27$ & $98.11\%$ \\
        \bottomrule
         
    \end{tabular}
    }
    \captionsetup{font=scriptsize}
    \vspace{-1em}
    \caption{Experimental results w/o ground truth on BEHAVE. \textbf{Bold} denotes the result reported in main paper.}
    \captionsetup{font=normalsize}
    \label{tab:exp_wogt}
    
\end{minipage}
\hfill
\begin{minipage}[t]{0.333\linewidth}
\centering
    \resizebox{1.0\linewidth}{!}{
    \begin{tabular}{l | l | c}
        \toprule
         & & \multicolumn{1}{c}{Obj.} \\
        SMPL & Contact & mIoU $\uparrow$\\
        \hline\hline
        Multi-HMR & - & $74.80\%$ \\
        Multi-HMR & DECO & $75.02\%$ \\
        Multi-HMR & VLM & $77.64\%$ \\
        \textbf{GT} & \textbf{GT} & $80.15\%$ \\
        \bottomrule
    \end{tabular}
    }
    \captionsetup{font=scriptsize}
    \vspace{-1em}
    \caption{Different contact estimation methods.}
    \captionsetup{font=normalsize}
    \label{tab:diff_contact}
\end{minipage}
\vspace{-1.5em}
\end{table}

\subsection{Additional Details and Analysis on in-the-wild}\label{sec:human_completion}
\cref{fig:supple_pipeline} presents a pipeline without ground-truth annotations. \Cref{tab:exp_wogt} reports human mesh recovery accuracy in terms of MPJPE on the BEHAVE dataset, along with amodal completion results using predicted SMPL models and a Vision-Language Model for contact estimation.
Notably, Multi-HMR~\cite{baradel2024multi} shows a MPJPE less than 70mm and achieves performance comparable to ground truth annotations in both object and human completion. Multi-HMR proves to be robust in occluded environments.
We also illustrate the SMPL estimation results in \cref{fig:multihmr_overlay}.
\vspace{-1em}
\noindent\paragraph{Binary Contact Map.}
To improve practicality, we introduce a pipeline that does not rely on GT annotations. Although we discuss existing contact estimation methods (e.g., DECO \cite{tripathi2023deco}) in \cref{sec:conclusion}, these methods often fail to detect the presence of contact points, offering only marginal performance gains (see \cref{tab:diff_contact}). Hence, we illustrate a VLM-based pipeline in \cref{fig:complex_completion}.
A prompt-engineered VLM~\cite{liu2023llava, openai2024chatgpt4} takes an image and outputs both a textual description) Each ID corresponds to one point, and the estimated SMPL parameters then designate these joints as contact points. Similarly, for an image \cref{fig:multihmr_overlay}-(c) \textit{left}, the VLM will produce a description “a man is sitting on a chair" and the hips joint IDs. Conversely, for a description such as “a person stands in front of a table,” the VLM will not output any joint ID.
As a result, combining Multi-HMR~\cite{baradel2024multi} with VLM approach achieves performance comparable to GT annotations, with a 2.5\% gap as shown in \cref{tab:diff_contact}. We plan to release the pipeline w/o GT.
\vspace{-1em}
\noindent\paragraph{SMPL accuracy} 
Although imperfect SMPL estimation can cause challenges for object and human completion, \cref{fig:multihmr_overlay} and \cref{tab:diff_contact} show that current SOTA models generally provide robust SMPL parameters in HOI scenarios, yielding sufficiently accurate contact estimates for our method. We achieve an mIoU of 96.79\% for human completion. Even when SMPL parameters are misaligned due to occlusion, restricting the inpainting region to the intersection between the object segmentation mask and the projected human mesh effectively limits errors.

\begin{figure*}[t!]
    \centering
    \captionsetup{type=figure}
    \includegraphics[width=0.935\textwidth]{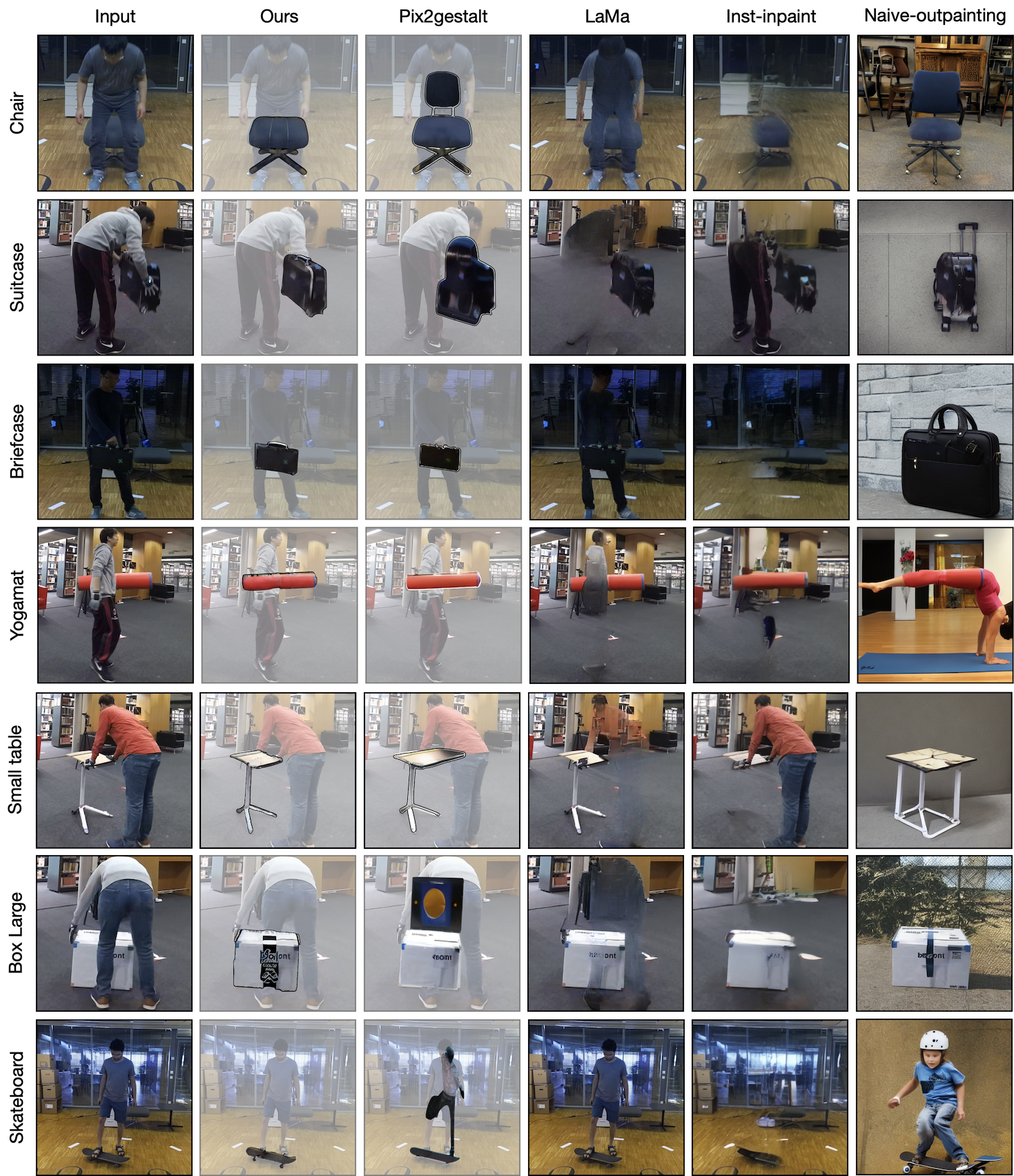}
    \captionof{figure}{Qualitative comparison between ours and baseline models.}
    \label{fig:qualitative_suppl}
\end{figure*}
\section{Additional Qualitative Results}
\subsection{Amodal Completion}
\paragraph{Baseline Comparison}
To illustrate the strengths of our method compared to existing approaches, additional results are provided in \cref{fig:qualitative_suppl}. These examples showcase our pipeline's ability to handle complex occlusion scenarios while preserving finer details. By comparison, baseline methods often fail to deliver coherent and detailed completions under similar conditions, underscoring the effectiveness of our approach.

\begin{figure*}[t!]
    \centering
    \captionsetup{type=figure}
    \includegraphics[width=1.0\textwidth]{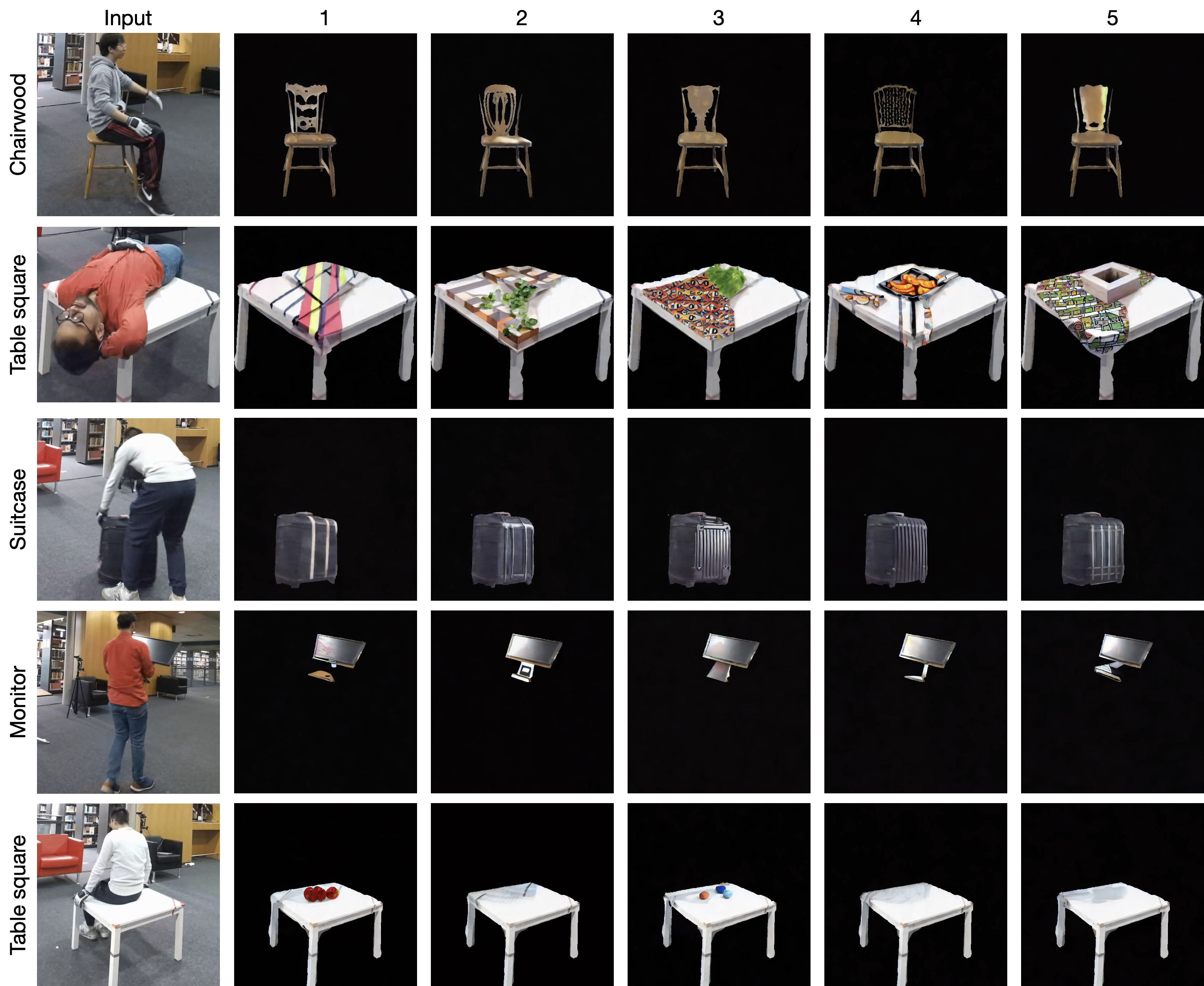}
    \vspace{-0.5em}
    \captionof{figure}{Diverse outputs generated by our pipeline. The visualization includes results from 5 different samples.}
    \label{fig:qualitative_suppl2}
\end{figure*}

\vspace{-1em}
\noindent\paragraph{Diverse Outputs}
The diverse outputs generated by our pipeline, visualized in \cref{fig:qualitative_suppl2}, highlight the flexibility of our approach in producing multiple plausible amodal completions for a single input. However, this diversity also exposes a limitation: the lack of consistency between these outputs. Addressing this challenge could drive future research, focusing on improving coherence across diverse completions to achieve more reliable and unified results, particularly for downstream tasks like 3D reconstruction.

\begin{figure*}[t!]
    \centering
    \captionsetup{type=figure}
    \includegraphics[width=0.95\textwidth]{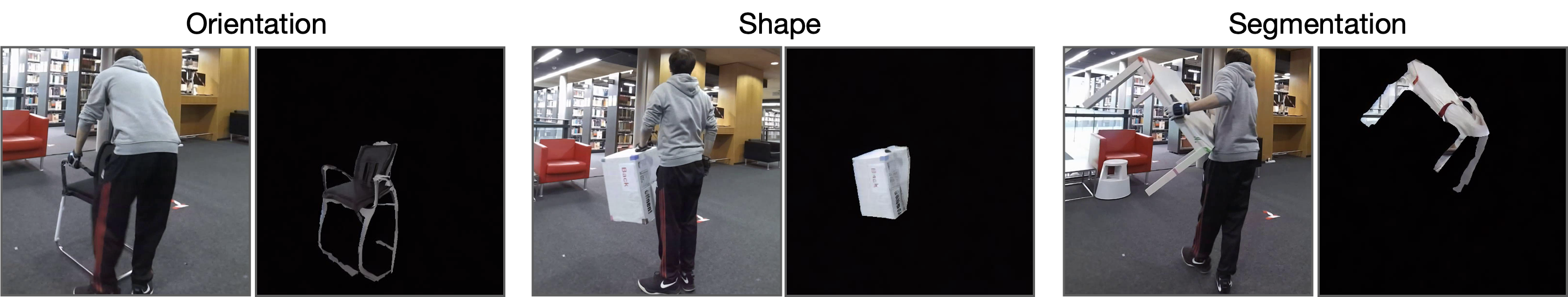}
    \captionof{figure}{Failure cases from our pipeline, categorized into orientation errors, shape errors, and errors caused by poor segmentation.}
    \label{fig:failure}
\end{figure*}

\vspace{-1em}
\noindent\paragraph{Failure Cases}
We visualize failure cases in \cref{fig:failure} to analyze the limitations of our approach, categorized into three types: 
1. \textit{Object Orientation Errors}: Misinterpreted object direction, often due to ambiguous visual cues, causes misalignment.
2. \textit{Shape Completion Errors}: Challenges in predicting occluded regions, especially for complex geometries, result in unrealistic shapes.
3. \textit{Segmentation Errors}: Inaccurate masks lead to flawed reconstructions, affecting amodal completion and 3D reconstruction.
Segmentation errors can be mitigated by user-driven manual corrections, while shape errors can be addressed by adjusting the parameter \( r \) in our pipeline. However, resolving orientation errors requires further research and is left as a direction for future work.

\subsection{3D Reconstruction}\label{suppl:3D_recon}
\begin{figure*}[t!]
    \centering
    \captionsetup{type=figure}
    \includegraphics[width=0.87\linewidth]{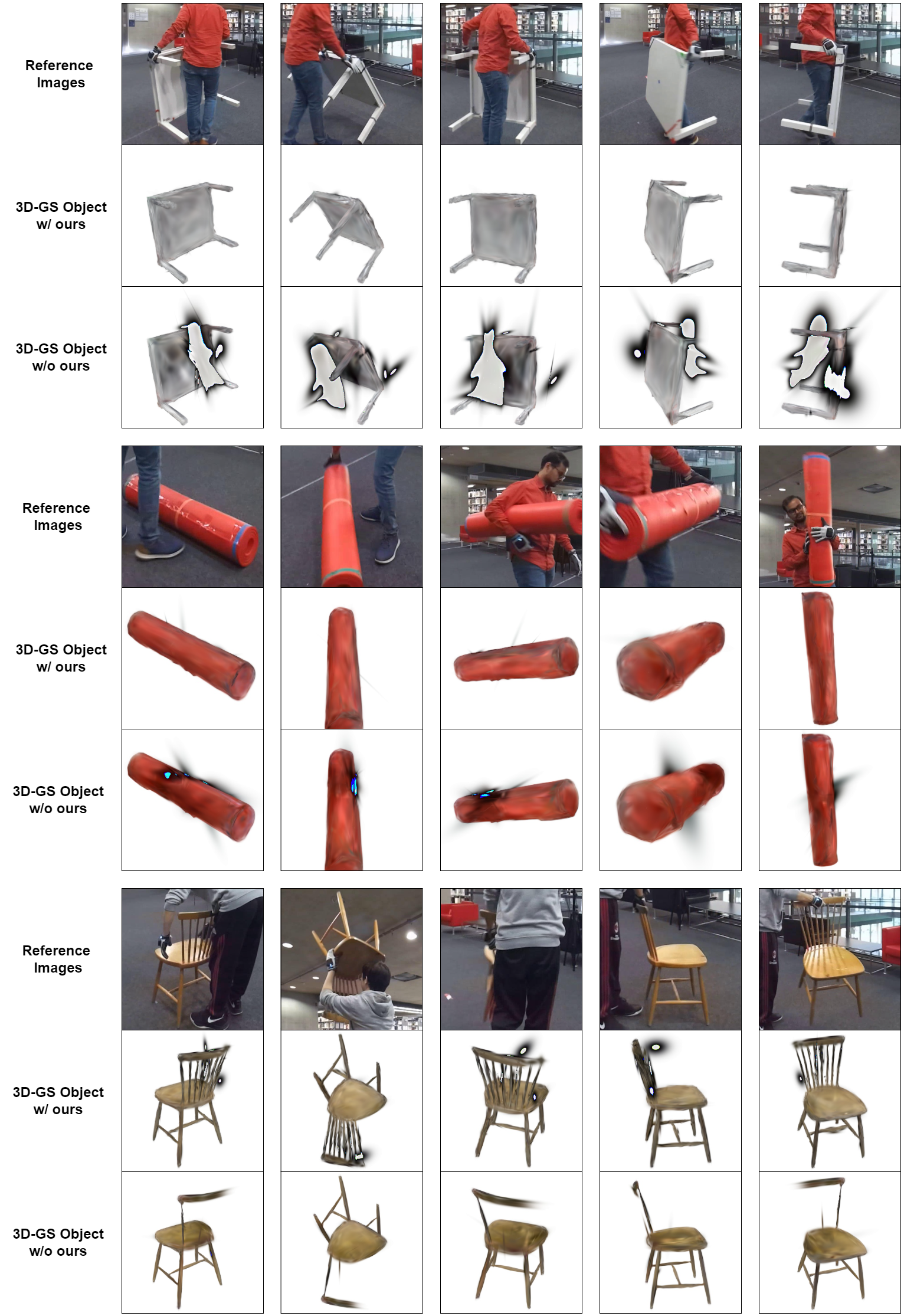}
    \captionof{figure}{Additional qualitative results of 3D-GS with and without amodal completion.}
    \label{fig:gs_obj_avatar_suppl}
\end{figure*}

The comparison of 3D reconstruction results in \cref{fig:gs_obj_avatar_suppl} highlights the effectiveness of using amodally completed images over original occluded images. These results demonstrate that our amodal completion method significantly enhances the quality of 3D reconstructions, validating its role as a vital preprocessing step for complex 3D tasks. Additionally, we provide videos showcasing novel-pose synthesis with human-object interaction in the attached file.

\subsection{User study}\label{suppl:User_study}
\begin{figure*}[t!]
    \centering
    \captionsetup{type=figure}
    \includegraphics[width=0.80\linewidth]{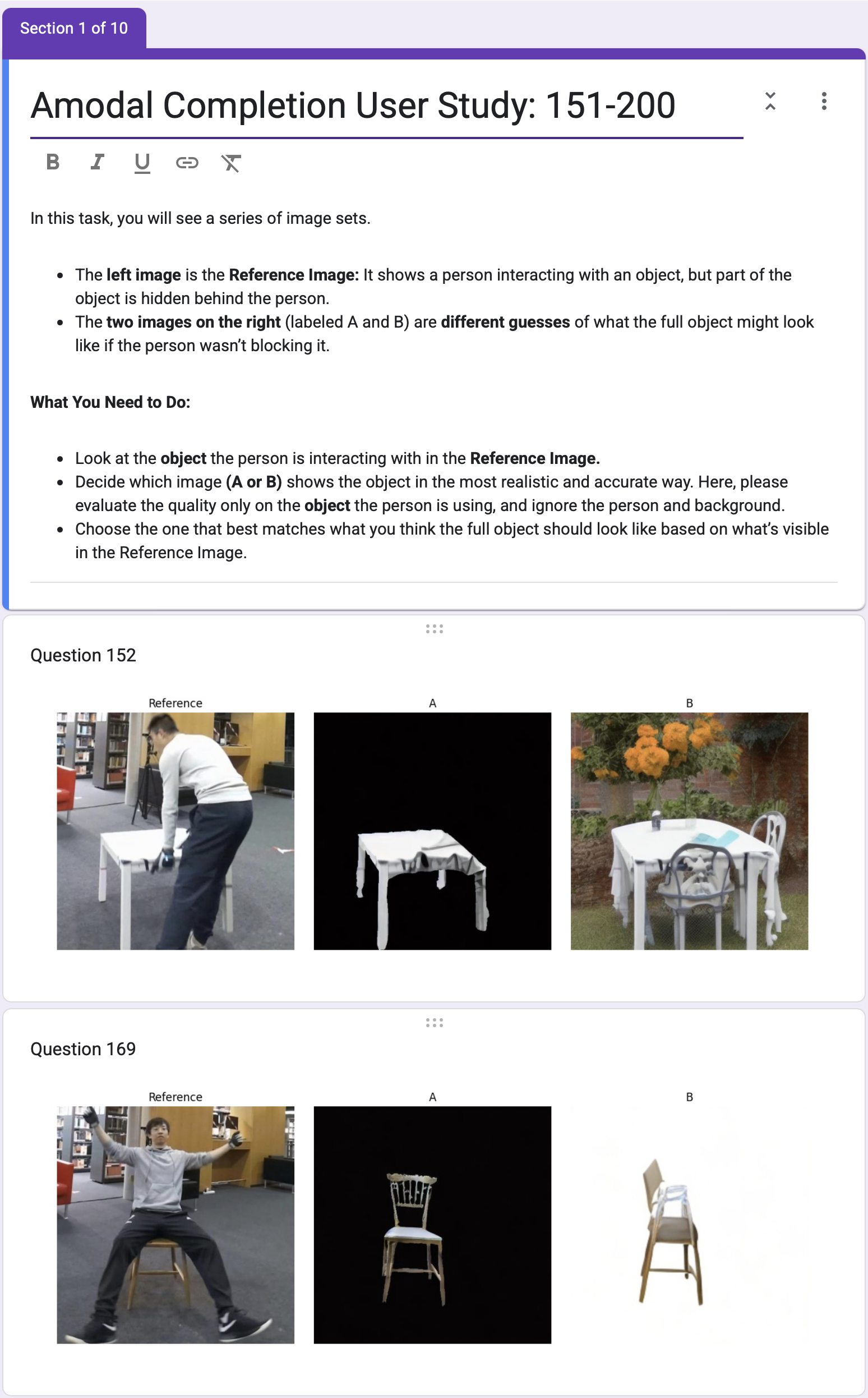}
    \captionof{figure}{User study instruction and examples.}
    \label{fig:userstudy}
\end{figure*}

Recognizing that CLIP score and mIoU have limitations in fully representing amodal completion quality, we conducted a user study. A total of 223 sample pairs were presented, with each pair evaluated by an average of 10 users. For each pair, users were asked to select the more accurate and realistic amodal completion result. This study focused exclusively on object amodal completion. Instructions and examples for the user study are provided in~\cref{fig:userstudy}.

\end{document}